\documentclass[fleqn,10pt, table, dvipsnames]{wlscirep}
\usepackage[utf8]{inputenc}
\usepackage[T1]{fontenc}
\usepackage{subfig} 
\usepackage{float} 
\usepackage{wrapfig}        
\usepackage{multirow}
\usepackage{amsmath,amssymb,amsfonts}
\usepackage{csquotes}
\usepackage[ruled,vlined]{algorithm2e}

\title{Detecting algorithmic bias in medical-AI models using conformal trees}

\author[1,*]{Jeffrey Smith}
\author[2]{Andre Holder}
\author[3]{Rishikesan Kamaleswaran}
\author[1]{Yao Xie}
\affil[1]{School of Industrial and Systems Engineering,
Georgia Institute of Technology, Atlanta, GA, 30332, USA}
\affil[2]{Department of Biomedical Informatics, Emory University School of Medicine, Atlanta, GA, 30303, USA}
\affil[3]{Department of Surgery, Duke University School of Medicine, Durham, NC 27708, USA}

\affil[*]{jsmith312@gatech.edu}


\begin{abstract}
With the growing prevalence of machine learning and artificial intelligence-based medical decision support systems, it is equally important to ensure that these systems provide patient outcomes in a fair and equitable fashion. This paper presents an innovative framework for detecting areas of algorithmic bias in medical-AI decision support systems. Our approach efficiently identifies potential biases in medical-AI models, specifically in the context of sepsis prediction, by employing the Classification and Regression Trees (CART) algorithm with conformity scores. We verify our methodology by conducting a series of synthetic data experiments, showcasing its ability to estimate areas of bias in controlled settings precisely. The effectiveness of the concept is further validated by experiments using electronic medical records from Grady Memorial Hospital in Atlanta, Georgia. These tests demonstrate the practical implementation of our strategy in a clinical environment, where it can function as a vital instrument for guaranteeing fairness and equity in AI-based medical decisions.
\end{abstract}
\begin{document}

\flushbottom
\maketitle
%
%
\thispagestyle{empty}


\section{Introduction}
\label{sec:introduction}

Machine learning (ML) and artificial intelligence (AI) technologies are becoming increasingly prevalent in critical decision-making processes in industries such as finance \cite{Ahmed2022ArtificialReview, Dixon2020MachinePractice}, education \cite{Kucak2018MachineTrends, Luan2021AEducation, Tiwari2023TheExperiences}, and criminal justice \cite{Broussard2023MachineSystem,Avila2020TheSystems, Chiao2019FairnessJustice}. As a result, the deployment of these technologies in such consequential domains has given rise to significant ethical considerations, particularly in terms of the influence of societal biases on model fairness. In medical applications, this bias has the potential to disproportionately affect particular patient subgroups and further amplify pre-existing disparities. The well documented exacerbation of existing disparities in healthcare data \cite{Obermeyer2019DissectingPopulations, Pencina2020PredictionApplication, Larson2016HowAlgorithm, Larrazabal2020GenderDiagnosis, Gianfrancesco2018PotentialData}, underscores the urgency of identifying these biases to ensure fair and equitable ML applications in this domain, especially for diverse and often underrepresented patient sub-populations.

Broadly, fairness can be grouped into three  categories: {\it individual} \cite{Dwork2012FairnessAwareness}, {\it group} \cite{Dwork2012FairnessAwareness}, and {\it causality-based}\cite{Kusner2017CounterfactualFairness}. Group fairness, as opposed to causality-based fairness and individual fairness, which both necessitate domain expertise to establish a just causal framework and aim for equality solely among comparable individuals, operates without presumption of knowledge and pursues equality across groups often framed in terms of one-dimensional protected attributes such as race, gender, or socio-economic status. 

While there has been much interest in group fairness measures \cite{Narayanan2018Tutorial:Politics}, researchers have noted their limitations. According to research by Castelnovo et al. \cite{Castelnovo2022ALandscape}, simply excluding protected features from the decision-making process does not inherently guarantee demographic parity, which is achieved when both protected and unprotected groups have equal probability of being assigned to the positive predicted class. Achieving demographic parity may involve using different treatment strategies for different groups in order to mitigate the impact of correlations between variables, a strategy that may be considered inequitable or counter-intuitive. Dwork et al.  \cite{Dwork2012FairnessAwareness} further expound on a ``catalogue of evils'' that highlight numerous ways the satisfaction of existing fairness definitions could prove ineffective in offering substantial fairness assurances.

Although a number of group fairness metrics have been developed recently \cite{Hardt2016EqualityLearning, Chouldechova2017FairInstruments, Feldman2015CertifyingImpact, Dwork2012FairnessAwareness, Kusner2017CounterfactualFairness, Narayanan2018Tutorial:Politics}, Dwork and Ilvento \cite{Dwork2019FairnessComposition} raise a notable issue that predictors may be adjusted in a way that they meet independent group fairness criteria, but their predictions contradict fairness at an interconnected subgroup level. This more nuanced case of group fairness spanning multiple subgroups is termed {\it intersectional group fairness} \cite{Crenshaw2018Demarginalizing1989}. Within this context, intersectionality posits that the interaction between multiple dimensions of identity may result in distinct and varying degrees of prejudice directed towards different potential subgroups \cite{Gohar2023AChallenges}. More abstractly, this problem may be connected to the concept of identifying ``fairness gerrymandering,'' \cite{Kearns2018PreventingFairness} where a classifier's results are deemed ``fair'' for each specific group (such as race, gender, insurance status, etc.), but significantly violate fairness when it comes to structured subgroups, such as specific combinations of protected features. 

In the healthcare domain, medical-AI decision support systems frequently function as black-box models, oftentimes providing limited insight into the structure of their training data, if any, as well as no visibility into the parameters used in model development. Developing effective and fair prediction models in this context poses unique difficulties, such as the potential absence of patient demographic representation in the training data and, in some instances, the complete absence of demographic information. The distinct challenges of healthcare data coupled with the intersectional group fairness contradictions could result in both inaccurate diagnoses and suboptimal interventions for certain structured subgroups.

In this paper, we address the challenge of detecting ``algorithmic bias'' in medical-AI models. These models utilize discrete time intervals for data organization (i.e., the 1-hour epoch structure we use that is normalized to ICU admission). They also include outcome prediction, with a defined prediction horizon. In particular, we present a novel framework utilizing a well-studied statistical approach, namely Classification and Regression Trees (CART) decision trees to detect regions of bias generated by a medical-AI model via uncertainty quantification. Moreover, this framework allows researchers and clinicians to evaluate the reliability of a prediction model, for a patient considering their individual characteristics. This methodology can be used on the output of any arbitrary prediction model to evaluate the effectiveness of the model in making accurate predictions for a specific patient and to assess whether the model should be applied to that type of patient. Our goal can be summarized as follows:
\begin{center}
{\it Using data, we aim to detect ``algorithmic bias'', via uncertainty quantification, generated by inferior algorithmic performance and directly identify structured subgroups, defined by various combinations of attributes, impacted by this bias.} 
\end{center}
%
%
The contributions of the work include:
\begin{itemize}
    \item We present a model-agnostic framework to systematically and rigorously detect biased regions through the retrospective analysis of results generated by medical-AI prediction algorithms. This method addresses gaps in current fairness evaluation methods that requires one to preselect groups in which bias is tested and paves the way for safer and more trustworthy medical-AI applications.
    \item Empirically, we evaluate the effectiveness of our technique in recognizing biased regions by conducting case studies using both synthetic and real data. Our findings demonstrate our ability to identify biased regions and gain insights into the characteristics that define these regions. 
\end{itemize}

\section{Related Works}
\subsection*{Group Fairness}
Several studies have addressed the challenges of group fairness by developing predictors that ensure fairness across numerous subgroups via ``fairness auditing.'' Kearns et al. \cite{Kearns2018PreventingFairness} propose a zero-sum game played between an ``Auditor'' and ``Learner'' to evaluate a predictor's fairness by minimizing error while adhering to specified fairness constraints. Separately, Herbert-Johnson et al. \cite{Hebert-Johnson2018Multicalibration:Masses} introduce a post-processing iterative boosting algorithm which combines all subgroups $c \in \mathcal{C}$, where $\mathcal{C}$ represents a class of subgroups, until the model is $\alpha$-calibration. Pastor, Alfaro, and Baralis \cite{Pastor2021IdentifyingClassification} examine subgroup bias by exploring the feature space through data mining techniques.

\subsection*{Tree-based Failure Mode Analysis}
Although decision trees may not be regarded as the most sophisticated method for failure mode analysis, they have the significant advantage of yielding results that are easily interpretable by humans. Consequently, decision trees have become increasingly prominent as a method for failure mode analysis. Chen et al. \cite{Chen2004FailureTrees} train decision trees to diagnose failures in large-scale data systems by classifying system requests as successful or failed. Singla et al. \cite{Singla2021UnderstandingExtraction} apply decision trees to identify and explain failure modes of deep neural networks, focusing on robustly extracted features. They evaluate performance using metrics such as Average Leaf Error Rate (ALER) and Base Error Rate (BER) to identify high-error clusters of labeled images.  Nushi, Kamar, and Horvitz \cite{Nushi2018TowardsFailure} employ decision trees as part of their hybrid human-machine  failure analysis approach, \textit{Pandora}, which similarly identifies failure clusters in high-error conditions. 

In contrast to these works, our approach detects ``algorithmic bias'' within structured subgroups beyond binary classification contexts. It avoids computationally intensive exhaustive searches of all possible attribute combinations, integrates statistical rigor in the determination of bias, and does not explicitly rely on common fairness metrics which require the pre-selection of protected features.
\section{Preliminaries}
\subsection{Classification and Regression Trees (CART)}
\label{sec:cart}

Decision trees are a versatile and intuitive machine learning (ML) algorithm used for both classification and regression tasks, embodying a tree-link model of decisions and their possible consequences. The CART model \cite{Breiman2017ClassificationTrees}, is a non-parametric ML decision tree methodology that is well suited for the prediction of dependent variables through the utilization of both categorical and continuous predictors. CART models offer a versatile approach to defining the conditional distribution of a response variable $y$ based on a set of predictor values $x$ \cite{Chipman1998BayesianSearch}. 

\begin{figure}[htbp]
    \centering
    \includegraphics[width=.9\textwidth]{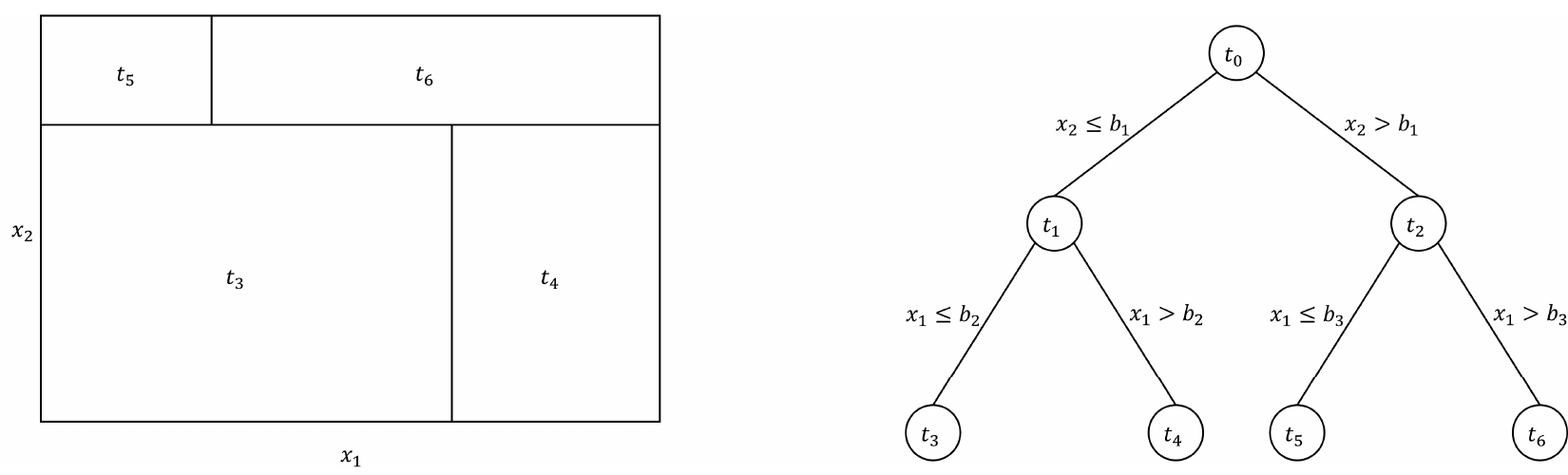}
    \caption{Example of an optimal axis-aligned decision tree with a depth of $K=2$ with $p=2$ dimensions. Splits occur along specific features in the form $x_j = b$ for $j = 1,2$.}
    \label{fig:cart_structure}
\end{figure}

In the classification setting, we are given the training data ($\mathbf{X,Y}$), containing $n$ observations ($\mathbf{x}_i, y_i$), $i = 1,...,n$, each with $p$ features $\mathbf{x}_i \in \mathbb{R}^p$ and a class label $y_i \in \{1,...,K\}$ indicating which of $K$ possible labels is assigned to this given point. In the regression setting our output variable is a continuous response variable $y_i \in \mathbb R$. Decision tree methods seek to recursively partition the dataset (feature space) into a number of hierarchically disjoint subsets with the aim of achieving progressively more homogeneous distributions of the response variable $y$ within each subset. An example of a decision tree is shown in Fig. \ref{fig:cart_structure}. 
 Beginning from the root node, an optimal feature and split point are identified based on an appropriate optimization metric. The feature, split-point pair defines the partition splitting the feature space, and this procedure is repeated for every sub-feature space that is created. These partitions will ultimately result in the binary tree structure consisting of interconnected root, branch, and leaf nodes.
\begin{itemize}
    \item {\it Root} nodes encapsulate the entire dataset, forming the foundational layer of the decision tree.
    \item {\it Branch} nodes are points in the dataset characterized by features and split points that serve as points of division for partitioning the feature space. Each of these branches extend to subsequent child nodes.
    \item {\it Leaf} nodes are the final nodes in the tree, classifying or predicting data points based on their localized patterns. 
\end{itemize}
%

CART models take a top-down approach and can be used for both classification and regression problems, as the name implies. Partitions are determined by using a specified loss function to evaluate the quality of a potential split and are based on both the features and values, that provide  optimal splits. The splitting criteria determine the optimal splits. In the classification setting, the criteria are often determined by the label impurity of data points within a partition. The splitting criteria for regression-based CART models focuses on minimizing the variance of data points in partitioned regions. CART models, as applied to both tasks, have two main stages: the decision tree's generation and subsequent pruning. We now transition to a more granular discussion on CART's implementation for both classification and regression problems.

\subsection*{Classification Trees}
\label{sec:classificationtrees}

The CART method, in the context of classification tasks, is a powerful tool for categorizing outcomes into distinct classes based on input features. The objective is to partition the feature space into regions that maximize the the uniformity of the response variable's classes within in each subsequent node during the partitioning process. This process begins at the root node and splits the feature space recursively based on a set of decision rules that maximally separate the classes.

When we consider splitting a classification tree, $T$, at any node $t$, we evaluate potential splits based on how well they separate the different classes of the response variable. For a given variable $X$, a split point $s$ is chosen to divide node $t$ into left ($t_L$) and right ($t_R$) child nodes. This division is based on whether the values of $X$ are less than or equal to $s$ or greater than $s$, formally defined as $t_L = \{{\textbf{X} \in t : X \leq s}\}$ and $t_R = \{{\textbf{X} \in t: X > s}\}$. The effectiveness of a split is measured using the impurity metric of Information Gain, which gauges the value of the insight a feature offers about a response variable. In practical applications, this measure is determined using Entropy or the Gini index.
\begin{itemize}
\item {\it Entropy} functions as a metric of disorder or unpredictability. It measures the impurity or randomness of a node, especially in binary classification problems. Mathematically, it is expressed as:
    \[ E = -\sum_{i=1}^{K} p_i \log_2 p_i,\]
where $p_i$ is the probability of an instance belonging to the $i^{th}$ class.

\item {\it Gini index} serves as an alternate measure of node impurity. Considered a computationally efficient alternative to entropy, it is formulated as follows:
\[ E = \sum_{i=1}^{K} p_{i} (1-p_{i}),\]
where, yet again, $p_i$ is the probability of an instance belonging to the $i^{th}$ class.

\item {\it Information Gain} is a metric calculated by observing the impurity of a node before and after a split and is formulated as:
\[\text{IG} = E_{{\rm parent}} - \sum_{i=1}^{K} w_{i}  E_{{\rm child}_{i}},\]
where $w_{i}$ is the relative weight of the child node with respect to the parent node.
\end{itemize}
The algorithm uses these splitting criteria to divide the feature space into sub-regions recursively, terminating when any of the specified stopping criteria are satisfied. After the dividing procedure finishes, each region gets assigned a class label $1,..., K$. This assigned class label will predict the classification of any points inside the region. Typically, the assigned class will be the most common class among the points in the region. 

\subsection*{Regression Trees}

Regression trees exhibit notable performance in the prediction of continuous output variables. The key aspect of their approach involves partitioning the feature space in such a way that the variation of the target variable is minimized within each segment of the space, referred to as nodes. To elaborate, when a regression tree, denoted as $T$, undergoes a split at a node $t$, we consider a potential division point, or split point $s$, for a variable $X$. This split point categorizes the data into left ($t_L$) and right ($t_R$) child nodes based on the condition whether $X \leq s$ or $X > s$. These nodes are formally represented as $t_L = \{{\textbf{X} \in t : X \leq s}\}$ and $t_R = \{{\textbf{X} \in t: X > s}\}$. The criterion for assessing the quality of a split in regression trees revolves around the variance within a node, given by
    \[
    \widehat{\Delta}(t) = \widehat{\rm VAR}(y|\textbf{X} \in t) = \frac{1}{n(t)} \sum_{\textbf{x}_i \in t} \left( y_i - \bar{y}_t \right)^2,
    \]
where $\bar{y}_t$ is the mean value of the target variable for the data points within node $t$ and $n(t)$ represents the count of these data points. The variance within the child nodes, left ($t_L$) and right ($t_R$), is similarly calculated. The decision to split a parent node $t$ into child nodes is based on the split that yields the highest decrease in variance, defined as 
\[
\widehat{\Delta}(s,t) = \widehat{\Delta}(t) - (\widehat{W}(t_L)\widehat{\Delta}(t_L) + (\widehat{W}(t_R)\widehat{\Delta}(t_R)),
\]
where $\widehat{W}(t_L) = n(t_L)/n(t)$ and $\widehat{W}(t_R) = n(t_R)/n(t)$ denote the proportions of data points in $t$ allocated to $t_L$ and $t_R$, respectively.

The process of developing the tree $T$ is iterative, identifying the variable and split point that maximizes variance reduction. Similar to its classification counterpart, the recursive partitioning of the feature space aims at reducing variance with the ultimate goal of accurately estimating the conditional mean response $\mu(x)$, in the tree's terminal nodes. The predicted response for data points in node $t$ is the mean target variable value, $\bar{y}_t$, for those points.

Without limitations, the tree generation process of the CART algorithm will continue until each data point is represented by a single leaf node. This is often not recommended as fully growing a tree to maturity introduces the risk of overfitting. To counter this, the tree development process includes constraints such as minimal sample split, maximum tree depth, and cost-complexity pruning to fine-tune the tree's structure and fit. 


\subsection{Conformal Prediction}
Conformal prediction is a statistical framework where the aim is to quantify uncertainty in the predictions made by some arbitrary prediction algorithm by converting point-predictions into set-valued functions with coverage guarantees. Consider a training set $\{(X_i,Y_i)\}^n_{i=1}$ and a test point $\{X_{n+1},Y_{n+1}\}$ sampled i.i.d. from some unknown distribution $P$. Using $\{(X_i,Y_i)\}^n_{i=1} \cup \{X_{n+1}\}$ as input, conformal prediction produces a set-valued function, denoted by $\hat{C}(\dot)$, that satisfies the guarantee $\mathbb{P}(Y_{n+1} \in \hat{C}(X_{n+1})) \geq 1-\alpha$, where $\alpha \in (0,1)$ is a nominal error level.

\section{Conformal tree based method for algorithm bias detection}

Given a pre-trained prediction algorithm $\mathcal A$, our objective is two-fold. First, can we detect the presence of bias in the predictions made by the algorithm? Second, if bias is detected, can we precisely identify the region $\mathcal S$ within the $p$-dimensional feature space where the algorithm exhibits suboptimal performance, a region we term the ``algorithmic bias'' region. In this context, $p$ denotes the number of features which can be categorical and/or continuous valued. 

We assume that the true region $\mathcal{S}$ is defined by a subset of key variables (features) $j \in S$. For real-valued features, this is represented as $X_j \in [L_j, U_j]$, where $L_j$ and $U_j$ represent some lower and upper bounds, respectively. For categorical value features, $X_j \in C_j$, $j \in S$. For example, if $p = 10$ and $S = \{1, 3\}$, the algorithmic bias region might be defined by age $X_1 \in [35, 50]$ and gender $X_3 = \{{\rm Female}\}$.

\begin{wrapfigure}{r}{0.5\textwidth}
    \centering
    \includegraphics[width = .4\textwidth]{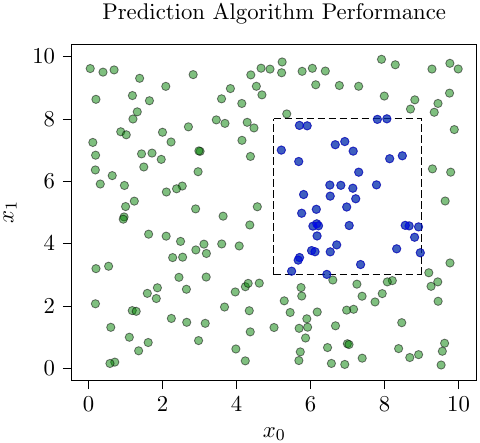}
    \caption{Illustration of the algorithmic bias region $\mathcal S$ in the feature space, where the algorithm $\mathcal A$ exhibits suboptimal performance.}
    \vspace{-.1in}
    \label{fig:ex_bias_region}
\end{wrapfigure}

This formulation implies that the subset of variables in the set $S$ will be the most critical in causing the bias, defining the algorithmic bias region $\mathcal S$. For instance, in our example, age and gender are the two most important features in defining the algorithmic bias region $\mathcal S$. Fig. \ref{fig:ex_bias_region} depicts the concept, where green dots signify superior performance, blue dots indicate worse performance, and the algorithmic bias region is delineated by a dashed-line box inside the feature space for $X \in \mathbb R^p$.

%
Without knowing the true algorithmic bias region, $\mathcal S$, of the algorithm $\mathcal A$, as represented using blue dots in Fig. \ref{fig:ex_bias_region}, we want to estimate it using test data. We can evaluate the performance of the algorithm on a collection of test samples $x_{i} \in \mathbb{R}^{p}$, $i= 1, \ldots, n$. The response associated with each test sample is $y_i \in \mathbb R$. Based on this, we can evaluate the algorithm performance using residuals.
\[
\epsilon_i = y_i - f(x_i), \quad i = 1, \ldots, n.
\]   
We note that alternative measures of algorithm performance, such as conformity scores, may replace residuals.

Our goal is to estimate the region $\widehat {\mathcal S}$ using $\{\epsilon_i\}_{i=1}^n$ as follows:
\begin{equation}
\widehat {\mathcal S} = \{X_j \in [L_j, U_j] \mbox{ or } X_j \in C_j, j \in \widehat S\},
\label{hat_S_define}
\end{equation}
where $S$, $L_j$, $U_j$, and $C_j$ are parameters to be determined.

Continuing with our previous example, if we estimate $\widehat S = \{1, 5\}$, this implies that we have correctly predicted the first important feature and incorrectly predicted the second. If $\widehat S = S$, then we estimate the correct subset variables used to define the algorithmic bias region. Once $S$ is estimated, the other parameters can be easier to decide. 

We hypothesize that the residuals within the bias region are larger. Thus, we formulate our problem as follows.
\begin{equation}
\label{opt_definition}
\begin{split}
\max_{\widehat S} \frac{1}{n(\widehat{S})} \sum_{x_i \in \widehat S} |\epsilon_i|,
\end{split}
\end{equation}
where $\widehat S$ is defined in \eqref{hat_S_define}, and $n(\widehat S)$ represents the number of data points $\widehat S$.

We apply decision trees, specifically Classification And Regression Trees (CART), as proposed by Breiman et al. \cite{Breiman2017ClassificationTrees}, to solve \eqref{opt_definition}. The CART algorithm recursively partitions the feature space until some stopping criteria are achieved and provides a piecewise constant approximation of the response function, here representing algorithm performance. The effectiveness of our methodology relies on the compactness of the estimated value $\widehat S$ to the true value $S$. 


\subsection*{Bias Testing}
Due to limited samples, bias estimation will have uncertainty, which we take into account in the bias detection through a conformal prediction procedure. This procedure provides a confidence interval for the estimated accuracy for each region. The confidence intervals are formed as follows. For each node in the decision tree, we can compute the confidence interval using the residuals $\epsilon_i$ of samples that fall into the region at a user-specified level $\alpha$, such that if bias exists, we detect it with at least probability $1-\alpha$. Confidence intervals are computed via quantiles. Formally defining $\text{Quantile}(\alpha; X) := \inf \{x: \alpha \leq \mathbb{P}(X\leq x)\}$, we obtain our lower and upper bounds via
\[
\hat{q}_{l} = \text{Quantile} \left( \frac{\alpha}{2}; \sum^{n}_{i=1}{\epsilon_i} \right), \quad 
\hat{q}_{u} = \text{Quantile} \left( 1- \frac{\alpha}{2}; \sum^{n}_{i=1}{\epsilon_i} \right)
\]
respectively, and confidence intervals via
\begin{equation}
    \hat{C}_{j}(x) = \left[\hat{f}_{j}(x) + \hat{q}_{l_{j}}, \hat{f}_{j}(x) + \hat{q}_{u_{j}}\right]
\end{equation}
where $\hat{f}_{j}(x)$ is the point prediction in the $j^{\text{th}}$ node of the decision tree.

\begin{figure}[!htbp]
    \centering
        \subfloat[]{%
            \includegraphics[width=.50\linewidth]{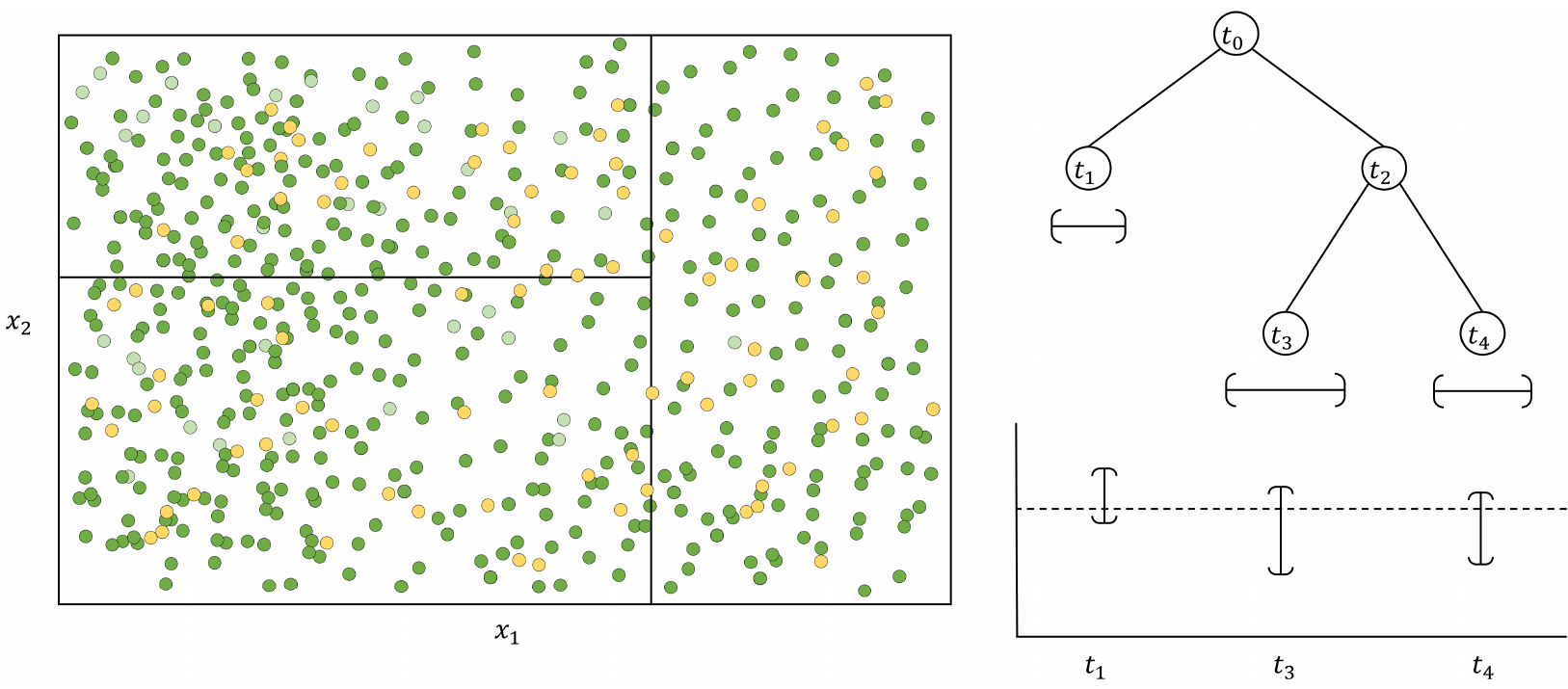}%
            \label{subfig:cp_nobias}%
        }
        \hfill
        \subfloat[]{%
            \includegraphics[width=.50\linewidth]{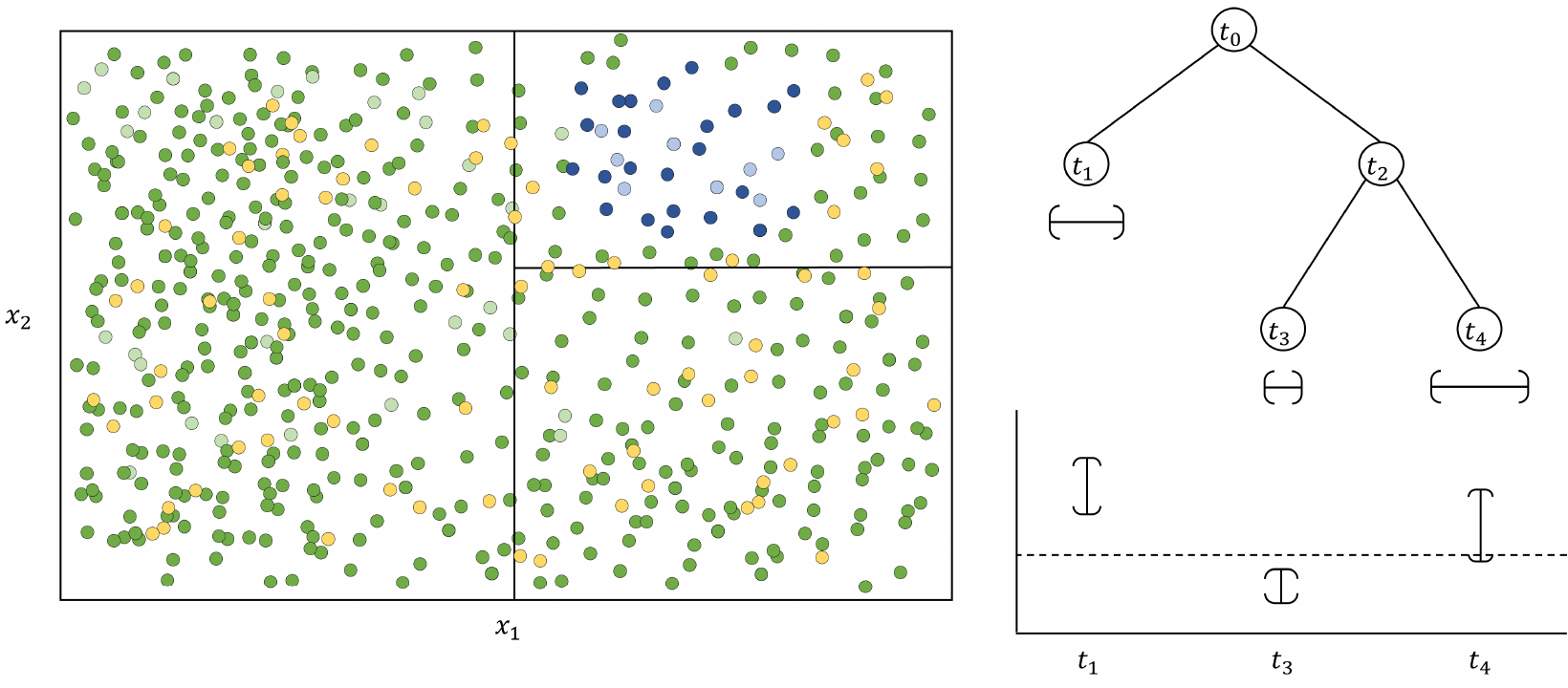}%
            \label{subfig:cp_bias}%
        }
        
    \caption{The plots present 2D examples of (a) the determination of no bias, and (b) the determination of bias when using the conformal prediction procedure within our bias detection framework.}
    \label{fig:cp_bias_example}
\end{figure}

To detect bias, we iterate over each terminal node, comparing the upper bound of the selected terminal node's confidence intervals with the lower bound of the remaining terminal nodes. When the confidence intervals mutually overlap, we can claim {\it no detection}, meaning that we believe that the node does not have sufficient statistical evidence to indicate that a particular group suffers from significantly larger bias. If the upper bound of the selected terminal node is less than or equal to the lower bound of the other terminal nodes, we consider that node to have bias at significance level $\alpha$. Alternatively stated, we are able to detect ``algorithmic bias'' with probability $1-\alpha$. Fig. \ref{fig:cp_bias_example} provides a visual example of the implementation of these confidence intervals in the bias detection procedure.  This bias detection method serves to audit the performance of any given pre-trained prediction algorithm $\mathcal A$ and is thus model agnostic.

\subsection*{Bias Detection Framework}

Let $D$ represent a dataset of patients, modeled as a tuple $(X, y)$, where $X \in \mathbb{R}^{m \times p}$ denotes a $p$-dimensional feature matrix for $m$ patients, and $y \in [0,1]$ represents the performance metric corresponding to the prediction outcome for each patient in the pre-trained prediction algorithm $\mathcal{A}(X)$. In this context, $X$ includes both categorical and continuous variables that capture the features of each patient, while $y$ evaluates the performance of the algorithm’s predictions on a scale from 0 (worst performance) to 1 (best performance).

Let $\alpha^*$ be the user-specified bias detection threshold, $K$ be the number of epochs, and $\Omega$ denote the hyperparameter space for the decision tree model. Our first objective is to identify a robust set of hyperparameters. For each epoch $k = 1, 2, \ldots, K$, we randomly shuffle the rows of the dataset $D$ and conduct a five-fold cross-validated grid search over the hyperparameter space $\Omega$, yielding the optimized set of hyperparameters $\Omega_k$.

Next, we fit our decision tree model $\Phi_k(D, \Omega_k)$ to the data. For each fitted decision tree $\Phi_k$, we test for the presence of bias at different nominal error levels $\alpha_i \in \{0.1, 0.2, \ldots, 0.9, 1.0\}$ using our conformal prediction procedure. If bias is detected at any nominal error level $\alpha_i \leq \alpha^*$, we conclude that bias is present at the user-specified threshold $\alpha^*$, otherwise the framework reports \textit{no bias}. We outline Algorithm~\ref{alg:bias_detection} below.

\begin{algorithm}
\caption{Bias Detection}
\label{alg:bias_detection}
\SetAlgoLined

\KwIn{Dataset $D = (X, y)$, Pre-trained prediction algorithm $\mathcal{A}(X)$, User-specified detection threshold $\alpha^*$, Number of epochs $K$, Hyperparameter space $\Omega$}
\KwOut{Bias detection result (Yes/No)}

\For{$k = 1$ \textbf{to} $K$}{
    Randomly shuffle the rows of dataset $D$\;
    
    Perform 5-fold cross-validated grid search over $\Omega$ to find optimized hyperparameters $\Omega_k$\;
    
    Fit decision tree model $\Phi_k(D, \Omega_k)$\;
    
    \For{each nominal error level $\alpha_i \in \{0.1, 0.2, \ldots, 1.0\}$}{
        Apply conformal prediction procedure to test for bias at $\alpha_i$\;
    }
}

\If{Bias is detected such that $\alpha_i \leq \alpha^*$ for any $\alpha_i$}{
    Report \textbf{Bias Detected}\;
}
\Else{
    Report \textbf{No Bias Detected}\;
}

\end{algorithm}

\section{Data}
In this section, we describe the dataset used in our real-world case study. We begin with a discussion of the sepsis definition and follow with the data pre-processing steps implemented prior to model development.
\subsection{Sepsis Definition}
We adopted the revised Sepsis-3 definition as proposed by Singer et al. \cite{Singer2016Thesepsis-3}, which defines sepsis as a life-threatening organ failure induced by a dysregulated host response to infection.  We implement the suspicion of infection criteria by identifying instances where the delivery of antibiotics in conjunction with orders for bacterial blood cultures occurred within a predetermined period. It is then determined that organ dysfunction has occurred when there is at least a two-point increase in the Sequential Organ Failure Assessment (SOFA) score during a specified period of time. The SOFA score is a numerical representation of the degradation of six organ systems (respiratory, coagulatory, liver, cardiovascular, renal, and neurologic) \cite{Jones2009ThePresentation}. This definition was utilized to identify patients meeting the sepsis criteria and to ascertain the most likely onset time of sepsis. 
\subsection{Cohorts}
\subsubsection{Grady Memorial Hospital}
Electronic health record (EHR) data was collected from 73,484 adult patients admitted to the intensive care unit (ICU) at Grady Memorial Hospital in Atlanta, Georgia from 2016 - 2020. This data included a total of 119,733 individual patient visits, referred to as ``encounters'', where, 18,464 (15.42\%) visits resulted in the retrospective diagnosis of sepsis. For our study, we excluded patients with less than 24 hours of continuous data, as well as, patients diagnosed with sepsis within the first six hours, reducing our dataset to 10,274 patient encounters involving 9,827 unique patients. Among these, 1,770 (17.23\%) visits were retrospectively diagnosed with sepsis during their ICU stay. The general demographic and clinical characteristics of the analyzed cohort of patients are summarized in Table \ref{table:grady_patients}.
\begin{table}[h!]
\caption{Baseline characteristics of Grady patients grouped by cohort.}
\centering
\resizebox{\columnwidth}{!}{
\begin{tabular}{ r r r r r r } \hline
\multicolumn{6}{r}{Grouped by sepsis}\\ \hline
{Variable} & & {Overall} & {Non-Sepsis} & {Sepsis} & {P-Value} \\
n & {} &             10274 &              8504 &              1770 &         \\
Age, median [Q1,Q3] &        &  53.0 [36.0,65.0] &  53.0 [36.0,64.0] &  54.0 [36.0,66.0] &   0.248 \\
Gender, n (\%) & Female &       3429 (33.4) &       2909 (34.2) &        520 (29.4) &  $<$0.001 \\
              & Male &       6845 (66.6) &       5595 (65.8) &       1250 (70.6) &         \\
Race, n (\%) & Asian &         125 (1.2) &          99 (1.2) &          26 (1.5) &  $<$0.001 \\
              & Black &       6711 (65.3) &       5631 (66.2) &       1080 (61.0) &         \\
              & Hispanic &         479 (4.7) &         387 (4.6) &          92 (5.2) &         \\
              & Other &         305 (3.0) &         233 (2.7) &          72 (4.1) &         \\
              & White &       2654 (25.8) &       2154 (25.3) &        500 (28.2) &         \\
ICU Length of stay (LOS), mean (SD) &        &         6.8 (9.4) &         4.3 (3.5) &       19.1 (16.5) &  $<$0.001 \\

LOS in hospital, mean (SD) &        &       14.7 (19.7) &       10.5 (10.5) &       34.7 (35.2) &  $<$0.001 \\

\hline
\end{tabular}
    }
\label{table:grady_patients}
\end{table}
\subsubsection{Emory University Hospital}
EHR data were collected from 580,172 adult patients admitted to the Emory University Hospital ICU in Atlanta, Georgia between 2013 and 2021. Of these visits, 67,200 (11.58\%) resulted in the retrospective diagnosis of sepsis. Following the same cohort generation procedure used for the Grady dataset, the Emory dataset was reduced to 69,232 patient encounters, of which 5,704 (8.24\%) were retrospectively diagnosed with sepsis during their ICU stay. The demographic and clinical characteristics of the Emory patient cohort are summarized in Table \ref{table:emory_patients}.

\begin{table}[ht!]
\caption{Baseline characteristics of Emory patients grouped by cohort.}
\centering
\resizebox{\columnwidth}{!}{
\begin{tabular}{ r r r r r r } \hline
\multicolumn{6}{r}{Grouped by sepsis}\\ \hline
{Variable} & & {Overall} & {Non-Sepsis} & {Sepsis} & {P-Value} \\
n &  & 69232 & 63528 & 5704 & \\

Age, median [Q1,Q3] &  & 63.0 [51.0,73.0] & 63.0 [51.0,73.0] & 63.0 [52.0,72.0] & 0.476 \\

\multirow[t]{2}{*}{Gender, n (\%)} & Female & 32141 (46.4) & 29596 (46.6) & 2545 (44.6) & 0.004 \\
 & Male & 37091 (53.6) & 33932 (53.4) & 3159 (55.4) &  \\

\multirow[t]{7}{*}{Race, n (\%)} & Asian & 1949 (2.8) & 1798 (2.8) & 151 (2.6) & <0.001 \\
 & Black & 27280 (39.4) & 24824 (39.1) & 2456 (43.1) &  \\
 & Multiple & 300 (0.4) & 270 (0.4) & 30 (0.5) &  \\
 & Other & 3751 (5.4) & 3344 (5.3) & 407 (7.1) &  \\
 & White & 35952 (51.9) & 33291 (52.4) & 2661 (46.7) &  \\

ICU Length of stay (LOS), mean (SD) &  & 6.3 (10.8) & 4.7 (8.1) & 16.1 (17.8) & <0.001 \\
LOS in hospital, mean (SD) &  & 12.6 (15.2) & 10.5 (11.7) & 25.9 (24.9) & <0.001 \\

\hline
\end{tabular}
    }
\label{table:emory_patients}
\end{table}

\section{Sepsis Prediction Model}
In developing the sepsis prediction model, we reference the model development procedure described in Yang et al. \cite{Yang2019EarlyOptimization}, which is one of the best-performing algorithms for sepsis detection. We detail the model development process in Appendix \ref{appendix:xgb_model}.

\section{Synthetic Data Experiments}
\label{sec:synthetic_exp}
In this section, we conduct experiments utilizing three synthetic data simulations using multidimensional uniform distributions. The objective of these simulations is to methodically assess the effectiveness of the conformal tree procedure in the context of detecting algorithmic bias regions. The first experiment evaluates the sensitivity of our bias detection approach when no bias exists. The final two experiments assess the effectiveness of the CART algorithm in the context of detecting algorithmic bias regions. This comparison is carried out by evaluating the coverage ratio, which serves as our primary performance criterion. This metric has been designed to effectively analyze and encompass the potential presence of an algorithmic bias region that may emerge within the feature space.

\subsection{Performance Metrics}
We introduce a refined performance metric, namely the coverage ratio, designed to account for the presence of distinct region(s) characterized by algorithmic bias within the feature space.

\subsubsection*{Coverage Ratio in n-Dimensional Space}
The Coverage Ratio ($\mathit{CVR}$) in $n$-dimensional space provides a measure of how well the estimated region approximates the true region in higher-dimensional space. The metric quantifies the relationship between the hypervolumes of the true and estimated regions compared to the overlapping hypervolume covered by both regions. When $n=2$ or $n=3$, $\mathit{CVR}$ is comparable to measuring the ratio of overlap between the area or volume of two sets, respectively. This metric is extended to higher-dimensional spaces as follows:

Given a dataset $\mathcal{D} \subset \mathbb{R}^n$, consider two n-dimensional bounded regions defined by sets $\mathcal{S}$ (true region) and $\mathcal{\hat{S}}$ (estimated region). Let $|S|$ and $|\hat{S}|$ represent the hypervolumes of the true and estimated regions, respectively, in the $n$-dimensional space, and let $|S \cap \hat{S}|$ denote the hypervolume of overlap common to both regions. Mathematically, we define $\mathit{CVR}$ as:
\begin{equation}
    \mathit{CVR} = \frac{1}{2} \left( \frac{|S \cap \hat{S}|}{|S|} + \frac{|S \cap \hat{S}|}{|\hat{S}|}\right).
\end{equation}

\subsection{Experiments}
In our first experiment, we evaluated the sensitivity of our method using synthetically generated datasets, without explicitly defining biased regions. We conducted 500 replications for each of the following sample sizes: $n_s = [500, 750, 1000, 2000,$ $ 3000, 6000, 8000]$, across dimensions $p \in [2, 3, 4, 5]$. The feature vectors $x_i$ for $i = 1, 2, \ldots, 5$ were drawn from a uniform distribution over the range $[-10, 10]$, and the corresponding $y$ values were generated from a uniform distribution $Y\sim U(0,1)$. 

We initialized the experiment by setting a significance level $\alpha = 0.2$, aiming to detect bias with a confidence level of $1 - \alpha = 0.80$. For each simulation run, we applied bootstrap aggregation (bagging) with five estimators, using majority voting to determine the presence of bias. The effectiveness of our bias detection framework was evaluated based on the false discovery rate.

In the subsequent experiments, we introduced a single implicit bias region across a variety of sample sizes and dimensions. We conducted 100 replications for each of the following sample sizes: $n_s = [150, 200, 300, 400, 500, 750, 1000, 2000]$, across dimensions $p \in [2, 3, 4]$. Similarly to the first experiment, the features $x_i$ for $i = 1, 2, \ldots, 4$ were sampled from a uniform distribution over the range $[-10, 10]$. 

To simulate an algorithmic bias region, we generated the corresponding $y$ values from a uniform distribution within the range [0.8, 1.0]. A central point, denoted $c_i$, was randomly selected within the feature space. Data points located within a defined distance from this central point were modified so that their corresponding $y$ values followed a uniform distribution within the interval $[0.3, 0.6]$. This region of reduced output values represents a potential area of algorithmic bias within the feature space.

The objective of the second experiment was to examine the relationship between the data sample topology and the performance of our bias detection framework when applied to a predefined algorithmic bias region. For each sample size, $n_s$, a single algorithmic bias region was established and consistently maintained across all replications as the benchmark (true region). The experiment focused on evaluating the positional variability of data points, where new data points were randomly generated in each replication. 

The primary goal of our third experiment was to assess how the location of the algorithmic bias region affects the performance of our detection framework. To isolate this effect, the topology of the feature space remained fixed across all replications, allowing us to focus on how variations in the bias region's location influence model performance. We evaluated the effectiveness of our bias detection framework using the Coverage Ratio ($\mathit{CVR}$) performance metric, which measures the alignment between the estimated region produced by the model and the predefined true bias region.

\subsection{Results}
\label{sec:sim_results}
Our simulations were designed with two primary objectives: first, to assess the framework's ability to detect bias in scenarios where no bias is present, and second, to explore the complex relationships between algorithmic bias regions and the topologies of the feature space. Table \ref{tab:falsediscovery_rate} presents the false discovery rates observed in the first experiment, where we tested the framework's sensitivity to bias detection in the absence of bias. The table shows results across various sample sizes ($n_s$) and feature space dimensionalities ($p$), where the findings indicate that false discovery rates decrease as sample sizes increase, with similar trends observed across different values of $p$. 

\begin{table}[htbp]
\begin{center}
\caption{False discovery rates across sample sizes and feature space dimensionalities.}
\begin{tabular}{cccccccc}

& \multicolumn{7}{c}{{Sample Size}}\\
\hline\textbf{$p$} & 500 & 750 & 1000 & 2000 & 3000 & 6000 & 8000 \\
\hline
2 & 0.0100 & 0.0040 & 0.0160 & 0.0120 & 0.0060 & 0.0080 & 0.0100 \\
3 & 0.0000 & 0.0025 & 0.0075 & 0.0000 & 0.0000 & 0.0000 & 0.0000 \\
4 & 0.0000 & 0.0060 & 0.0095 & 0.0149 & 0.0050 & 0.0000 & 0.0000 \\
5 & 0.0080 & 0.0040 & 0.0060 & 0.0087 & 0.0100 & 0.0050 & 0.0000 \\
\hline
\label{tab:falsediscovery_rate}
\end{tabular}
\end{center}
\end{table}

\begin{figure}
    \def\twidth{0.25}
    \centering
        \hspace*{\fill}%
        \subfloat[]{%
            \includegraphics[width=\twidth\linewidth]{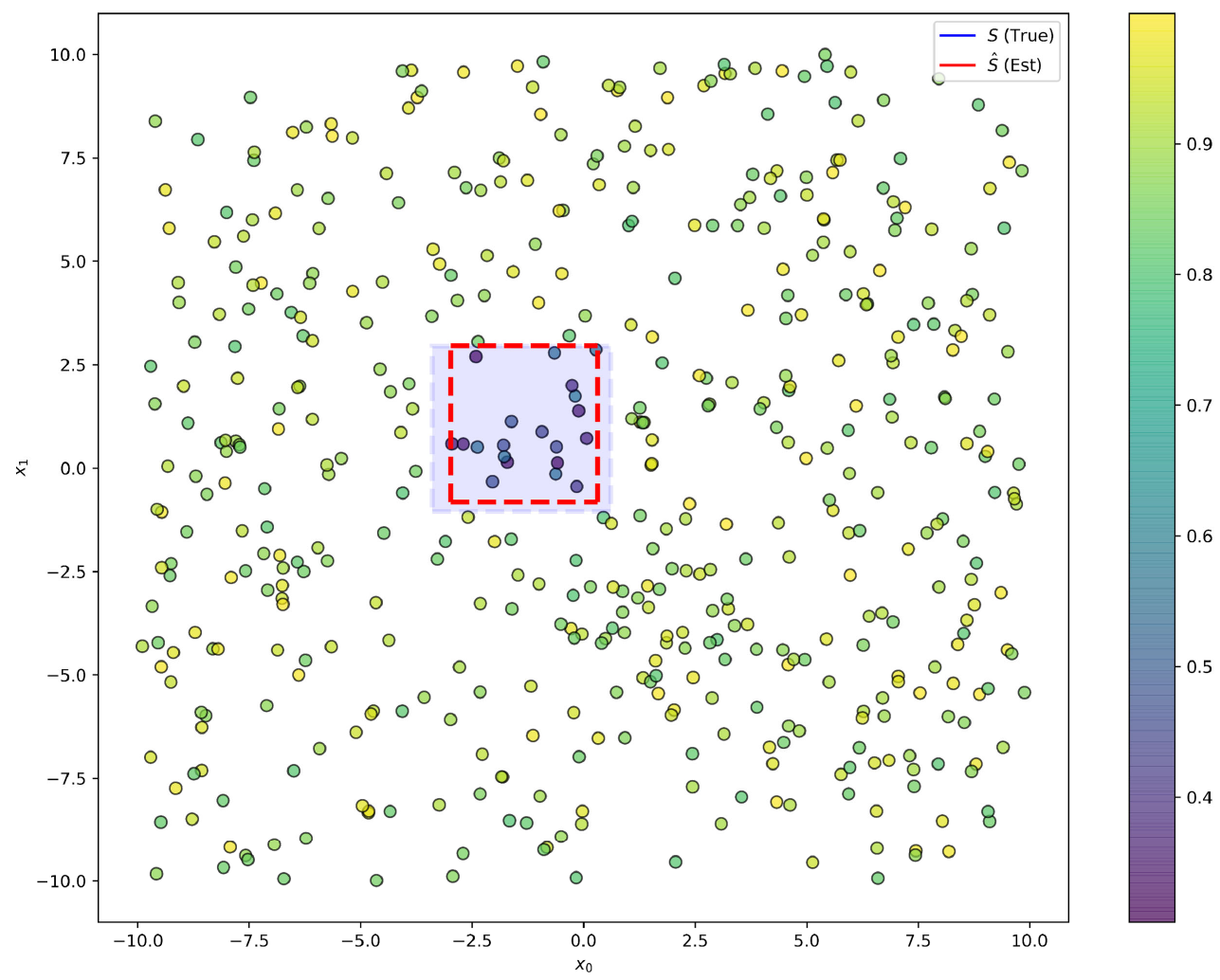}%
            \label{subfig:1}%
        }\hfill
        \subfloat[]{%
            \includegraphics[width=\twidth\linewidth]{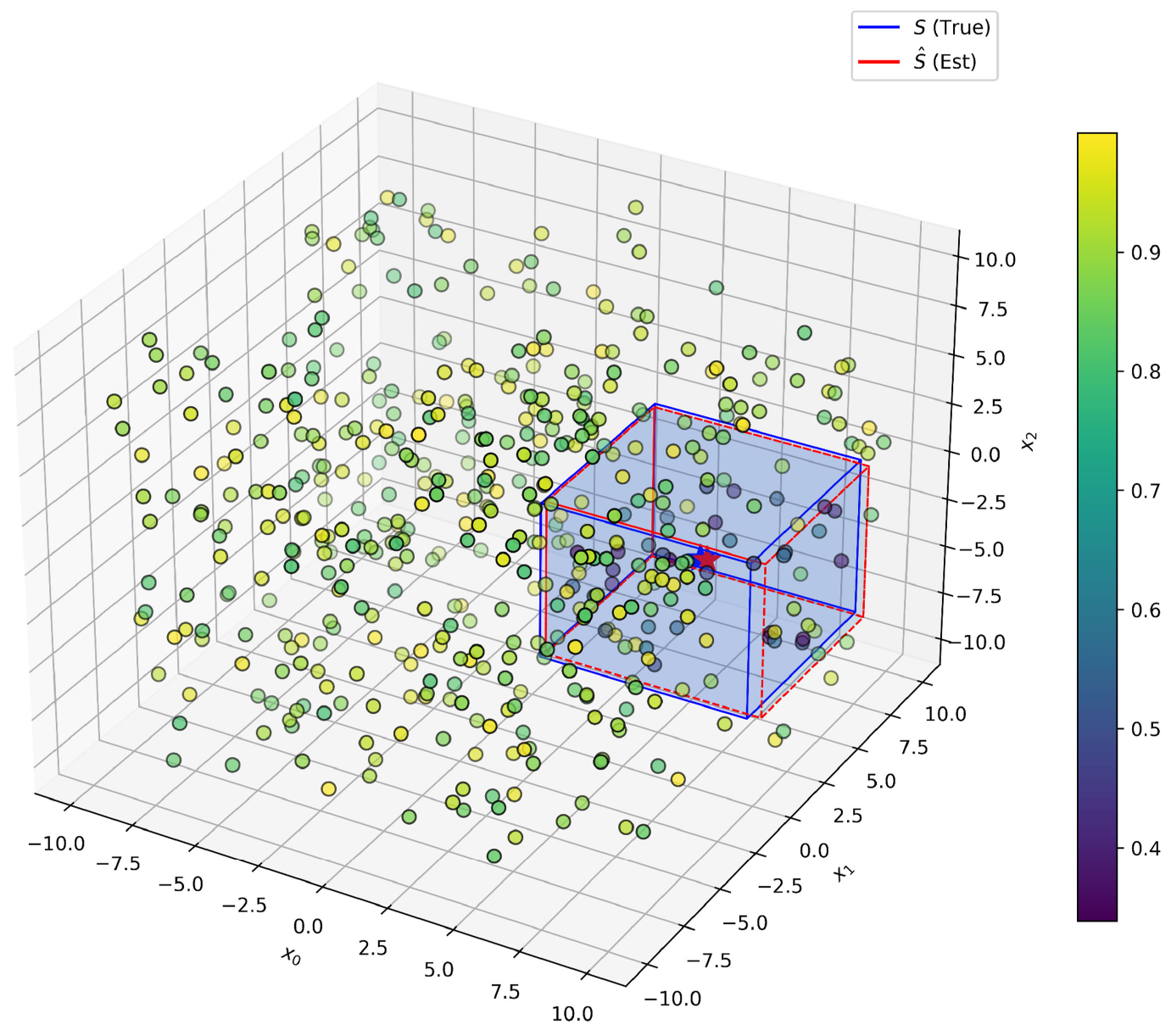}%
            \label{subfig:2}%
        }
        \hspace*{\fill}%
    \caption{Examples of the experimental results in 2(a) and 3(b) dimensional space.}
    \label{fig:bounded_sim}
\end{figure}

Fig. \ref{fig:bounded_sim} provides a visual representation of the ability of our approach to accurately estimate the borders of regions characterized by algorithmic bias. The true region(s) are delineated and filled in blue, whereas the estimated region(s) consist of points located inside the red dashed lines. Figs. \ref{subfig:1} and \ref{subfig:2} illustrate examples of the ability to identify bias regions in simulated output in the context of two and three-dimensional scenarios respectively.

\begin{figure}[htpb]
    \def\twidth{0.25}
    \centering
        \subfloat[]{%
            \includegraphics[width=\twidth\linewidth]{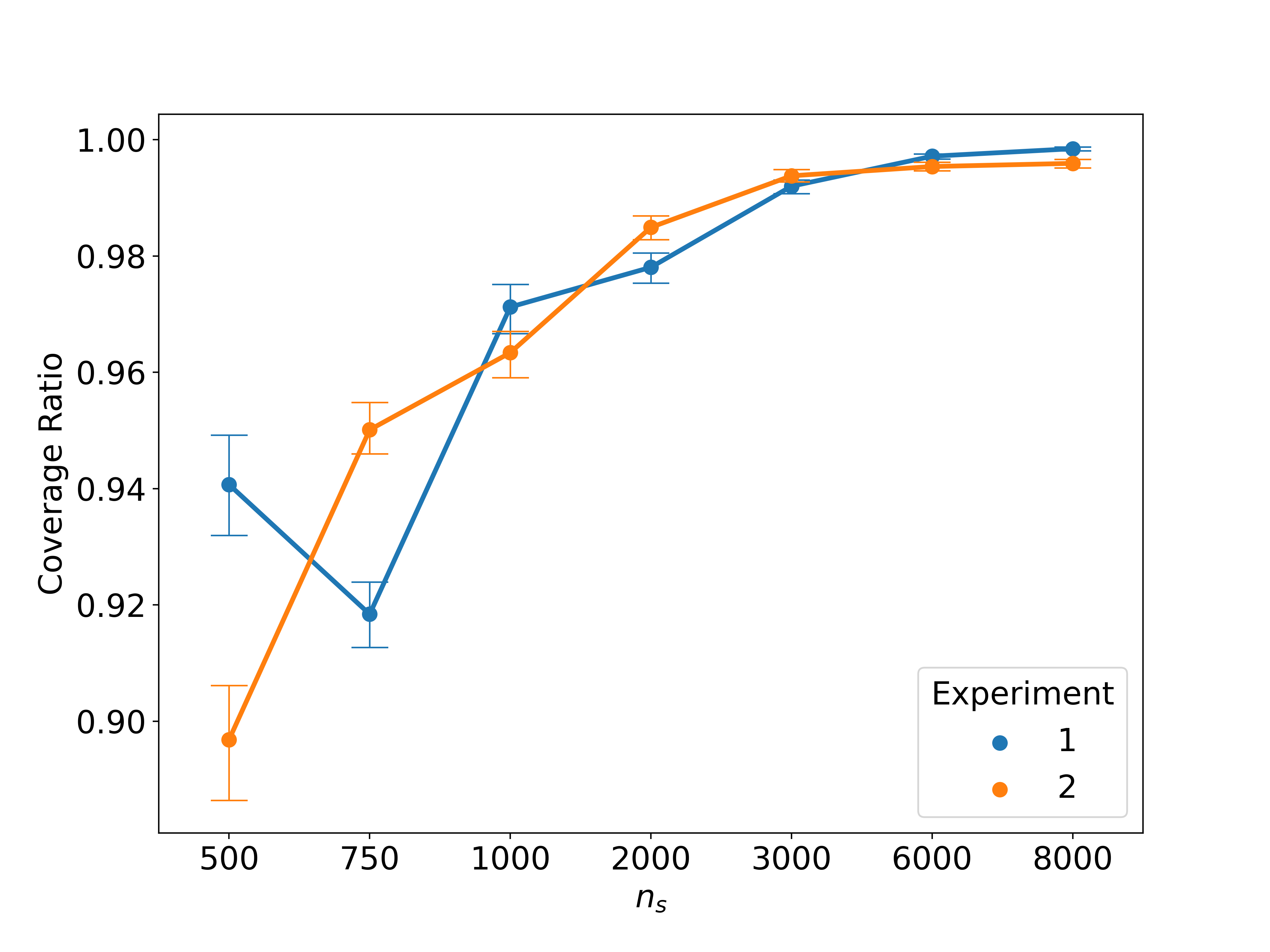}%
            \label{simResults_a}%
        }
        \subfloat[]{%
            \includegraphics[width=\twidth\linewidth]{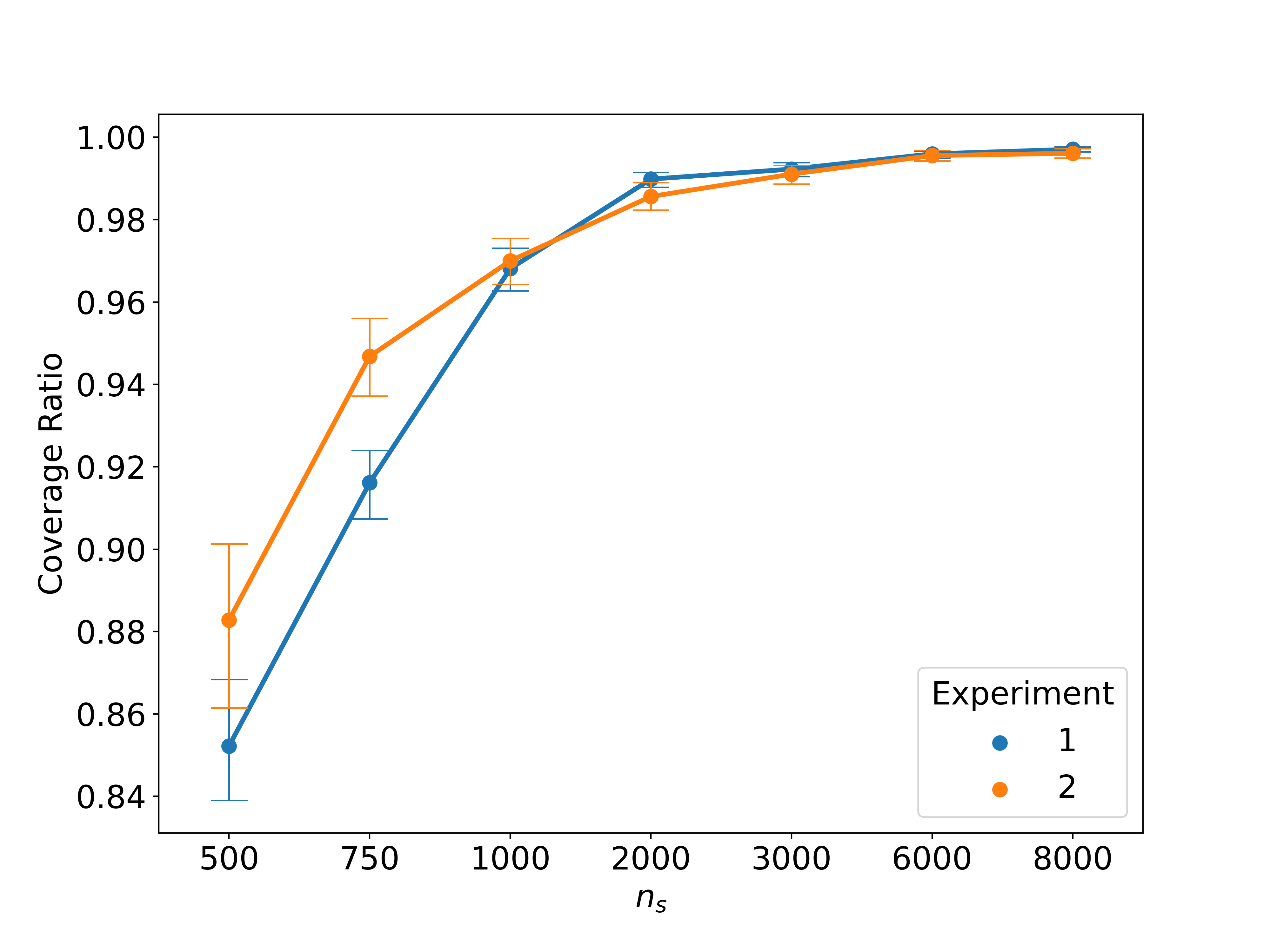}%
            \label{simResults_c}%
        }
        \hfill
        \subfloat[]{%
            \includegraphics[width=\twidth\linewidth]{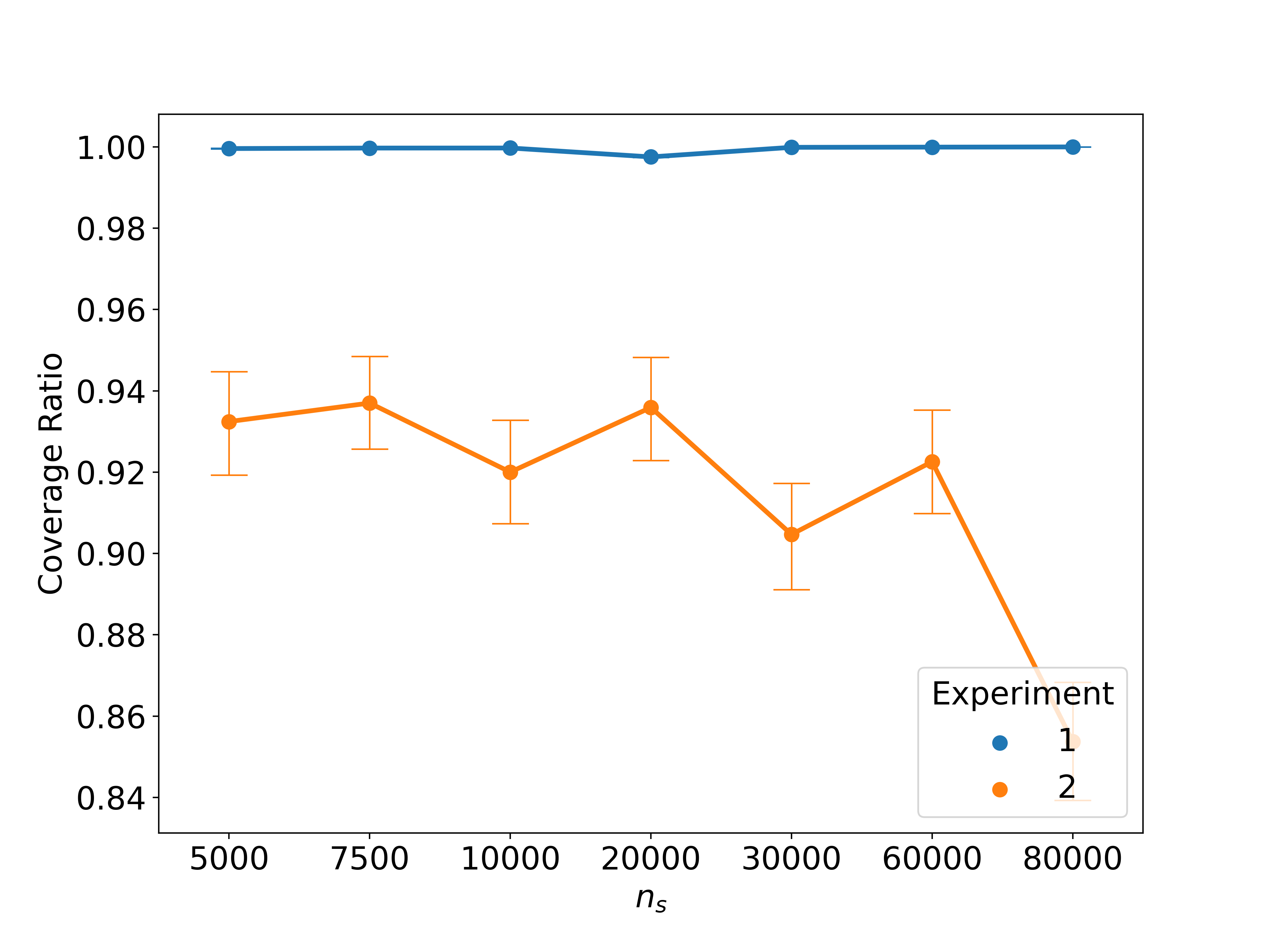}%
            \label{simResults_b}%
        }
        \subfloat[]{%
            \includegraphics[width=\twidth\linewidth]{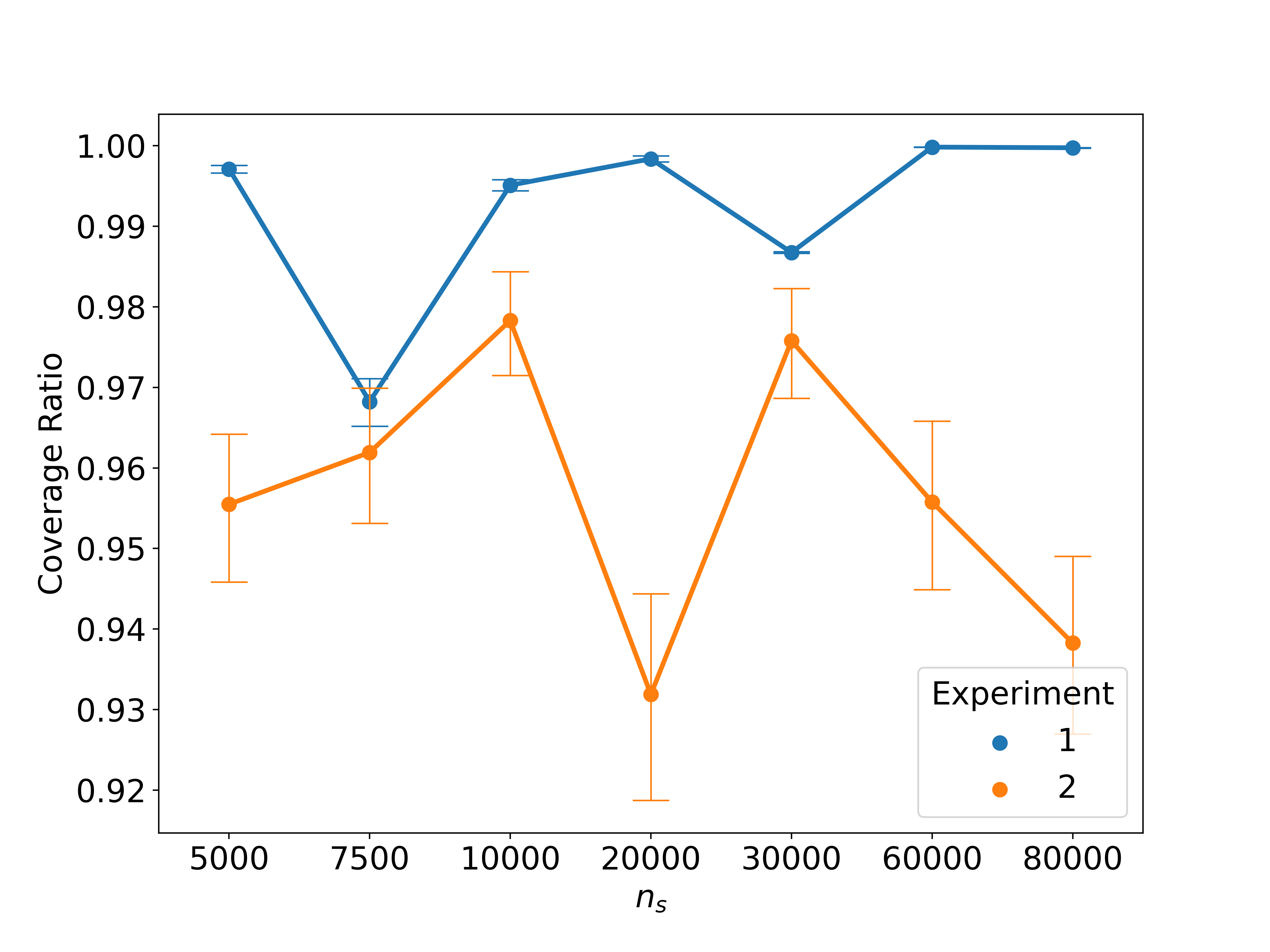}%
            \label{simResults_d}%
        }
    \caption{The plots show the mean coverage ratio for multiple $n$-dimensional test points: 2D(a), 3D(b), 4D(c), and 5D(d).}
    \label{fig:simResults}
\end{figure}

We provide a summary of the results achieved by our approach, as depicted in Fig. \ref{fig:simResults}, and confirm the efficacy of our bias detection framework in accurately detecting algorithmic bias regions. To provide precise details, Fig. \ref{fig:simResults} shows the mean performance of each experiment at the various sample size test points for multiple $n$-dimensional cases. The figures incorporate 95\% confidence intervals for both experiments. These results indicate that our method can efficiently detect the presence of 
 algorithmic bias layered in the feature space.
\section{Real-Data Experiment}

In the second phase of our empirical study, we evaluate the effectiveness of the sepsis prediction model and aim to identify any potential algorithmic biases. During this assessment, the test dataset is used to sequentially process the continuous data of each patient via the prediction model. We further refine the test data by only applying the model to patients whose EHR data includes at least one occurrence of sepsis. Implementing this approach results in an hourly forecast for every occurrence of a patient's data. Subsequently, we compute the performance of the classification model for every individual patient. Here, we selected model accuracy as the performance measure, implying it is the variable we are using to identify algorithmic bias. Next, we combine the accuracy of each patient's performance measure with their corresponding demographic data, which includes a range of factors such as gender, race, age, insurance type, and the existence and number of pre-existing comorbidities. One-hot encoding is used to transform non-numeric features into a numeric representation. Lastly, we define a threshold significance level $\alpha^* = 0.20$, meaning we want to detect ``algorithmic bias'' with a confidence level of at least $80\%$, and define our hyper-parameter space $\Omega$, as outlined in Table \ref{tab:hyperparams}. 
\begin{table}[htpb]
    \centering
    \caption{CART bias detection hyper-parameter tuning grid}
    \begin{tabular}{c|c}\hline
       Parameters  &  Grid\\
       \hline
        ``criterion'' & [``squared\_error'', ``absolute\_error'']\\
        ``splitter'' & [``best"]\\
        ``ccp\_alpha'' & [0.0, 0.0001, 0.0005, 0.001]\\
        ``max\_depth'' & [3,4]\\
        ``min\_samples\_leaf'' & [10, 30, 50, 60, 100]\\
        ``min\_samples\_split'' & [10, 30, 50, 60, 100]\\
        ``max\_features'' & [None,``log2'',``sqrt'']\\
        \hline
    \end{tabular}
    
    \label{tab:hyperparams}
\end{table}

\subsection{Results}
\subsubsection{Grady Memorial Hospital}
\begin{figure}[!htbp]
    \def\twidth{0.33}
    \centering
        \subfloat[]{%
            \includegraphics[width=\twidth\linewidth]{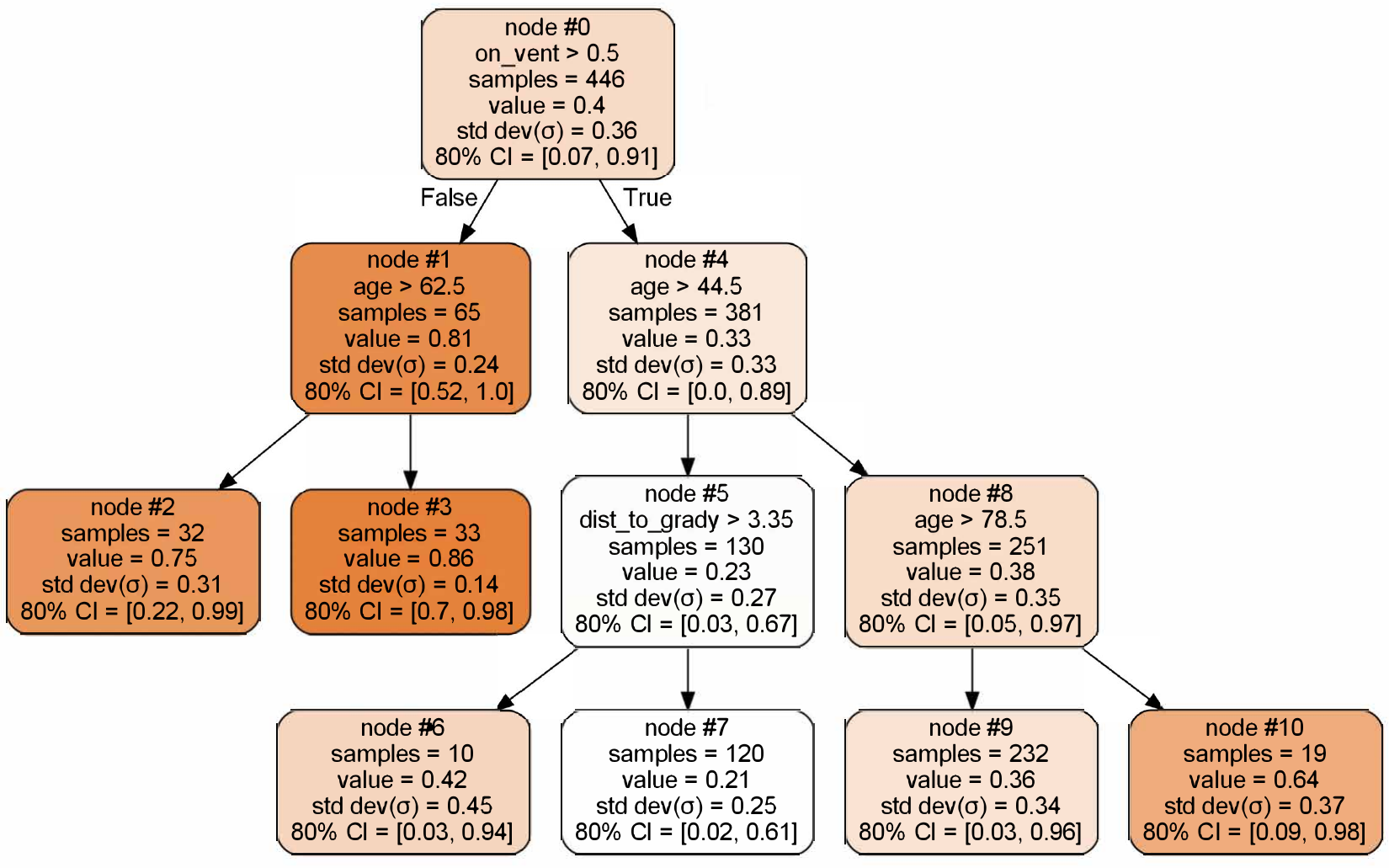}%
            \label{fig:biasresults_a}%
        }\hfill
        \hfill
        \subfloat[]{%
            \includegraphics[width=\twidth\linewidth]{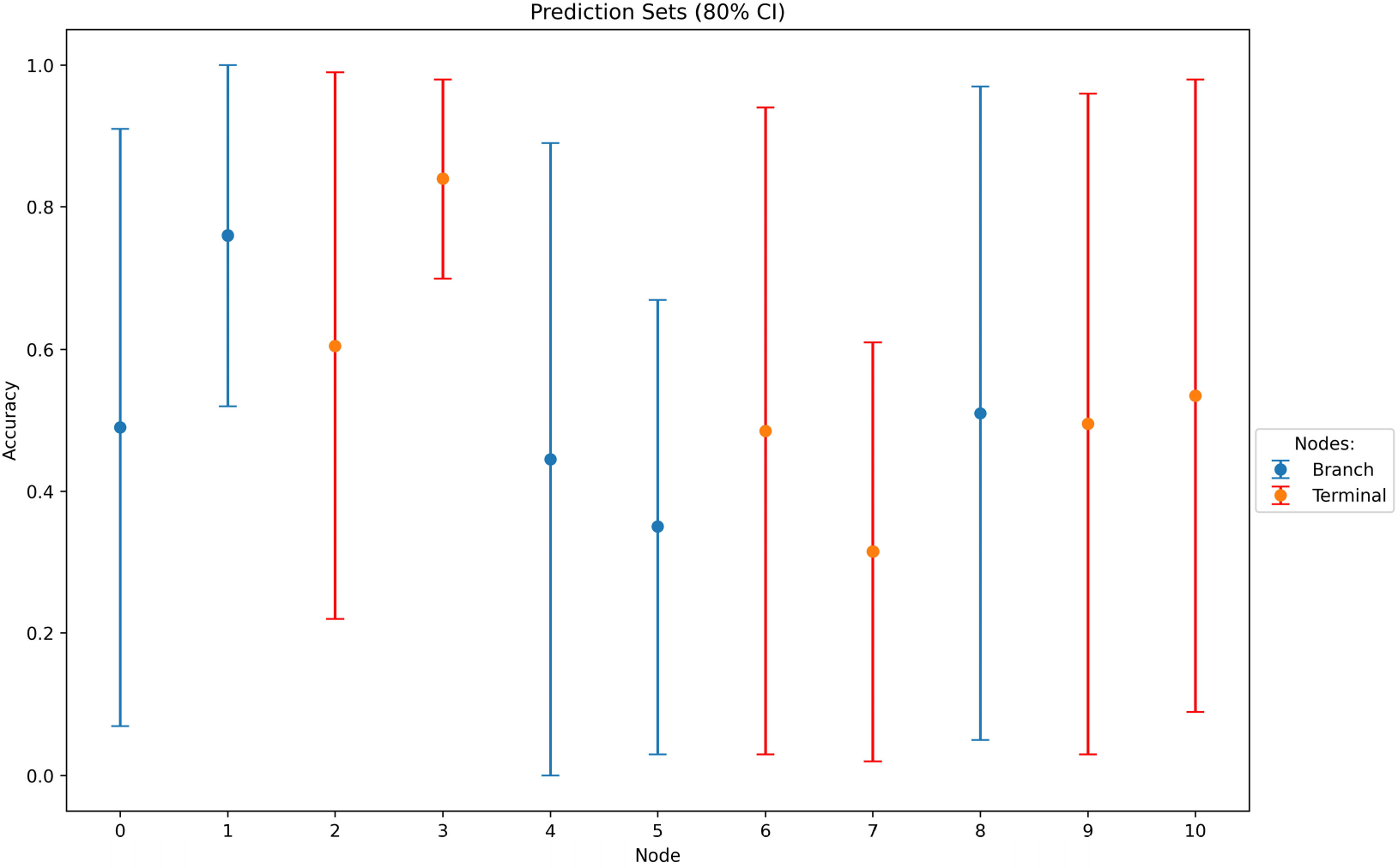}%
            \label{fig:biasresults_c}%
        }
        \hfill
        \subfloat[]{%
            \includegraphics[width=\twidth\linewidth]{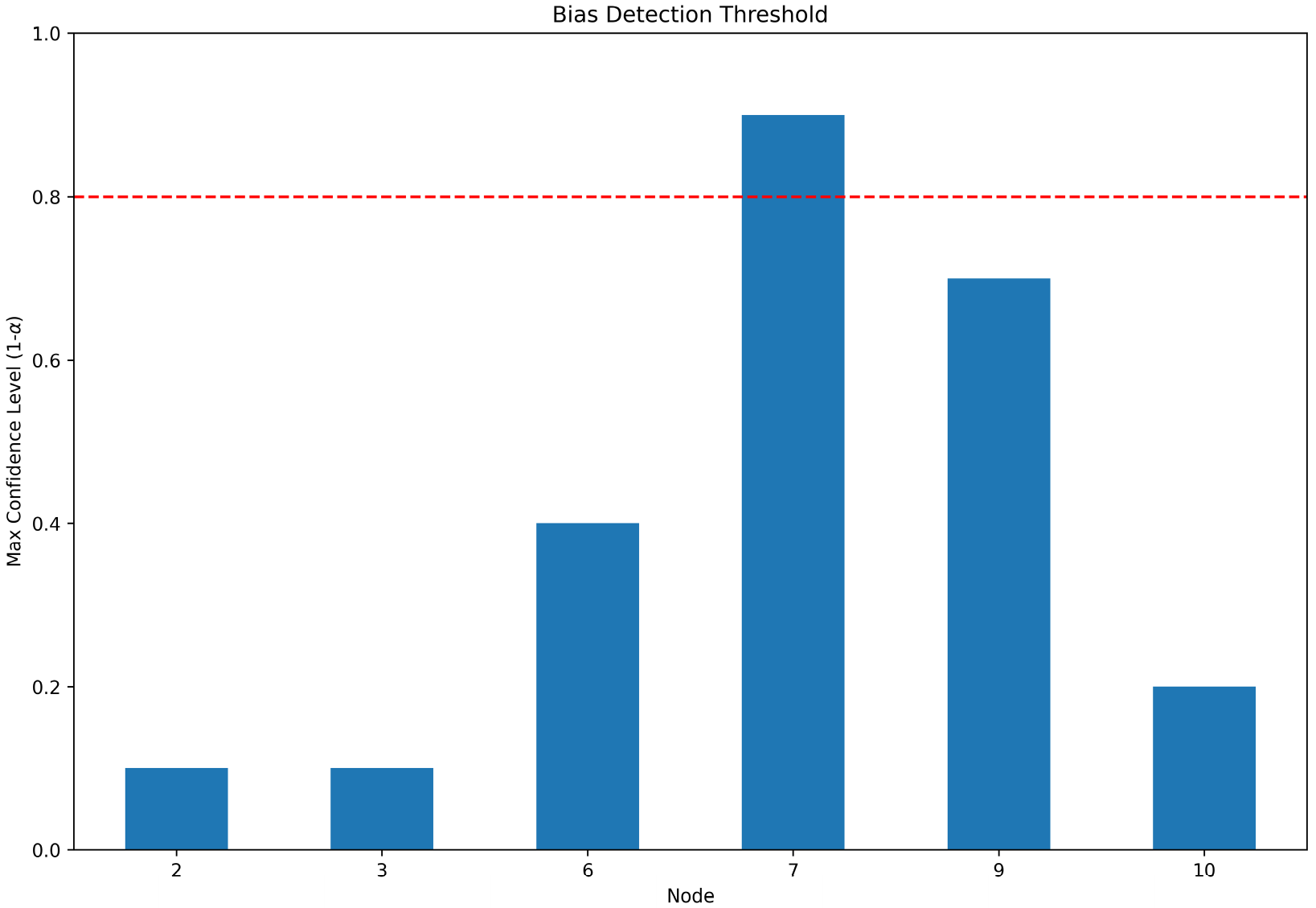}%
            \label{fig:biasresults_d}%
        }\hfill
        \subfloat[]{%
            \includegraphics[width=.50\linewidth]{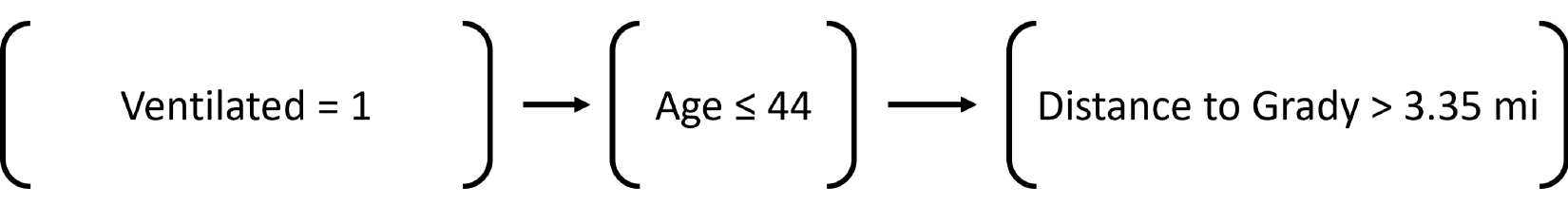}%
            \label{fig:biasresults_e}%
        }
    \caption{Grady bias detection model results. \ref{fig:biasresults_a} displays the complete decision tree, where the intensity of node shading corresponds to the magnitude of the point prediction—darker nodes indicate higher point prediction values, while lighter nodes indicate lower point prediction values. \ref{fig:biasresults_c} shows the predicted confidence intervals $\hat{\mathcal{C}}_{j}$ for each branch (blue) and terminal (red) node at significance level $\alpha$. \ref{fig:biasresults_d} presents the maximum bias detection confidence level $1-\alpha^{*}_{j}$ for the $j^{\text{th}}$ terminal node. \ref{fig:biasresults_e} provides a simplified representation of the nodes along that route.}
    \label{fig:biasresults}
\end{figure}

The final results of our bias detection framework for Grady Memorial Hospital are shown in Fig. \ref{fig:biasresults}. Our findings indicate that bias was detected for patients located in Node 7, with a significance level of $\alpha^* = 0.20$. Fig. \ref{fig:biasresults_a} illustrates the complete decision tree generated by the sepsis prediction model for the test dataset. Each node contains the feature split-point pair selected by the model at that node, the number of instances in the node, the predicted response variable $\hat{y}$ for the samples, the standard deviation within the node, and the conformal prediction set based on the significance level $\alpha^*$. 


Fig. \ref{fig:biasresults_c} visualizes the confidence intervals for each node's conformal predictions, providing a detailed view of prediction uncertainty across the tree. Fig. \ref{fig:biasresults_d} displays the optimized significance levels $\alpha^*$ across all leaf nodes, as summarized in Table \ref{tab:opt_alpha}. Notably, the optimized confidence level for Node 7 is 0.9, which translates to an optimized significance level of $\alpha_{7}^{*} = 0.10$.

Fig. \ref{fig:biasresults_e} provides a simplified representation of the key attributes that define this suboptimal path. Based on our bias detection analysis, we conclude that the sepsis prediction model $\mathcal{A}$ may underperform for the subgroup characterized as ``ventilated patients, younger than 45 years old, residing more than 3.35 miles from Grady Hospital.'' This summary not only highlights the algorithmic bias detected but also provides valuable insight into the demographic and clinical attributes associated with suboptimal model performance. 

\begin{table}[!ht]
 \caption{Optimized significance level $\alpha$ per node.}
    \label{tab:opt_alpha}
    \centering
   
    \begin{tabular}{c|c|c} 
        \hline
        Node & $\alpha^*$ & Confidence Level\\
        \hline
        2 & 1.0 & 0.00\\
        3 & 1.0 & 0.00\\
        6 & 0.60 & 0.40\\
        7 & \textbf{0.10} & \textbf{0.90}\\
        9 & 0.30 & 0.70\\
        10 & 0.80 & 0.20\\
        \hline
    \end{tabular}
\end{table}

\begin{figure}[!htbp]
    \def\twidth{.50}
    \centering
\includegraphics[width=\twidth\linewidth]{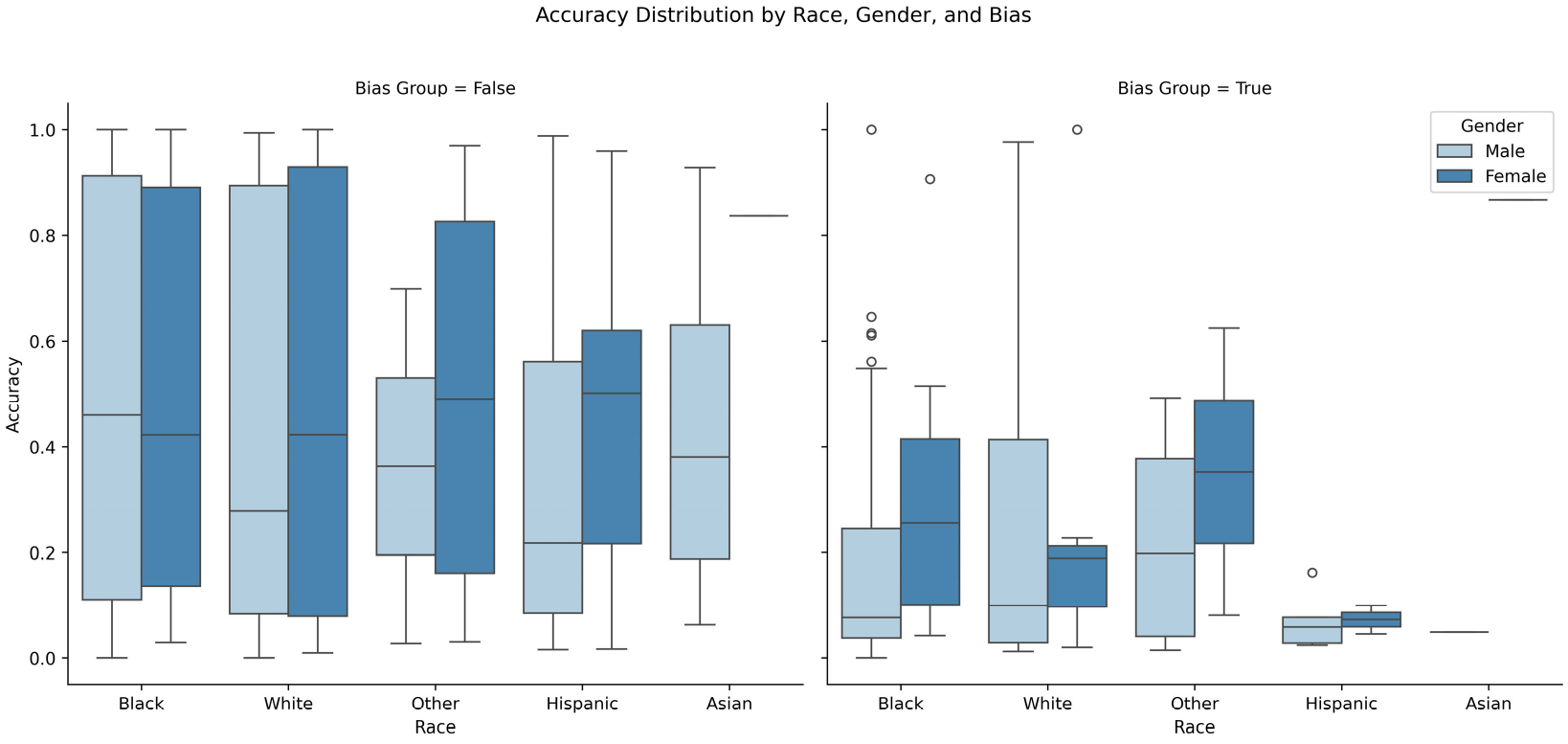}%
    \caption{Analysis of bias detection model results. This plot displays the distribution of accuracy scores grouped by Race, Gender, and Bias, highlighting differences in model performance across different sub-groups.
    }
    \label{fig:bias_analysis}
\end{figure}

Additionally, Fig. \ref{fig:bias_analysis} visualizes the distribution of accuracy scores defined by race, gender, and bias group. Each subplot represents a different bias category, with individual boxes for each combination of race and gender. This plot illustrates notable differences in the accuracy scores between patient subgroups based on bias group identification. Furthermore, it highlights gender-based differences within the biased group, showing that, on average, this model performs worse for men. 

\subsubsection{Emory University Hospital}

\begin{figure}[!htbp]
    \def\twidth{0.33}
    \centering
        \subfloat[]{%
            \includegraphics[width=\twidth\linewidth]{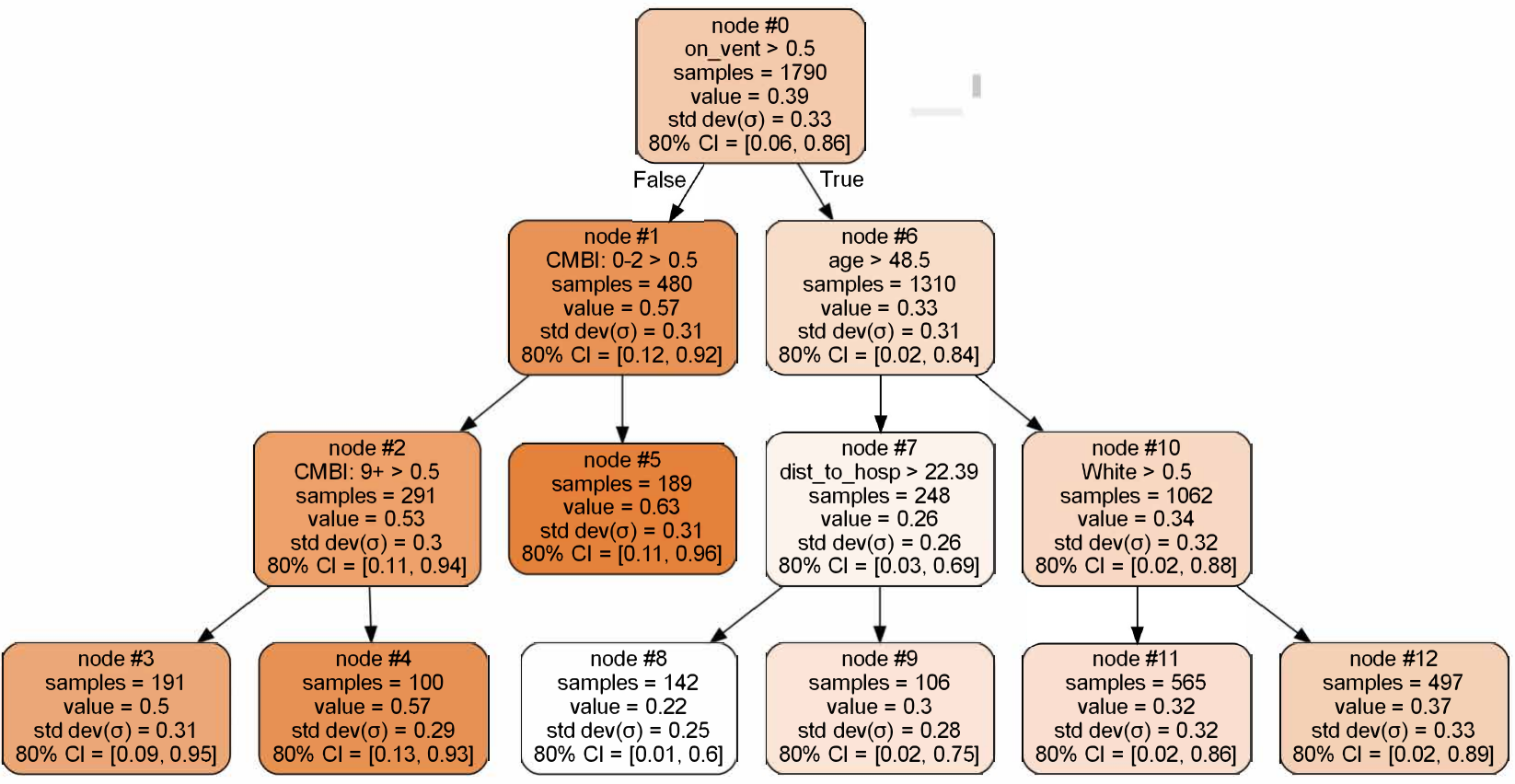}%
            \label{subfig:emory_biasresults_a}%
        }\hfill
        \subfloat[]{%
            \includegraphics[width=\twidth\linewidth]{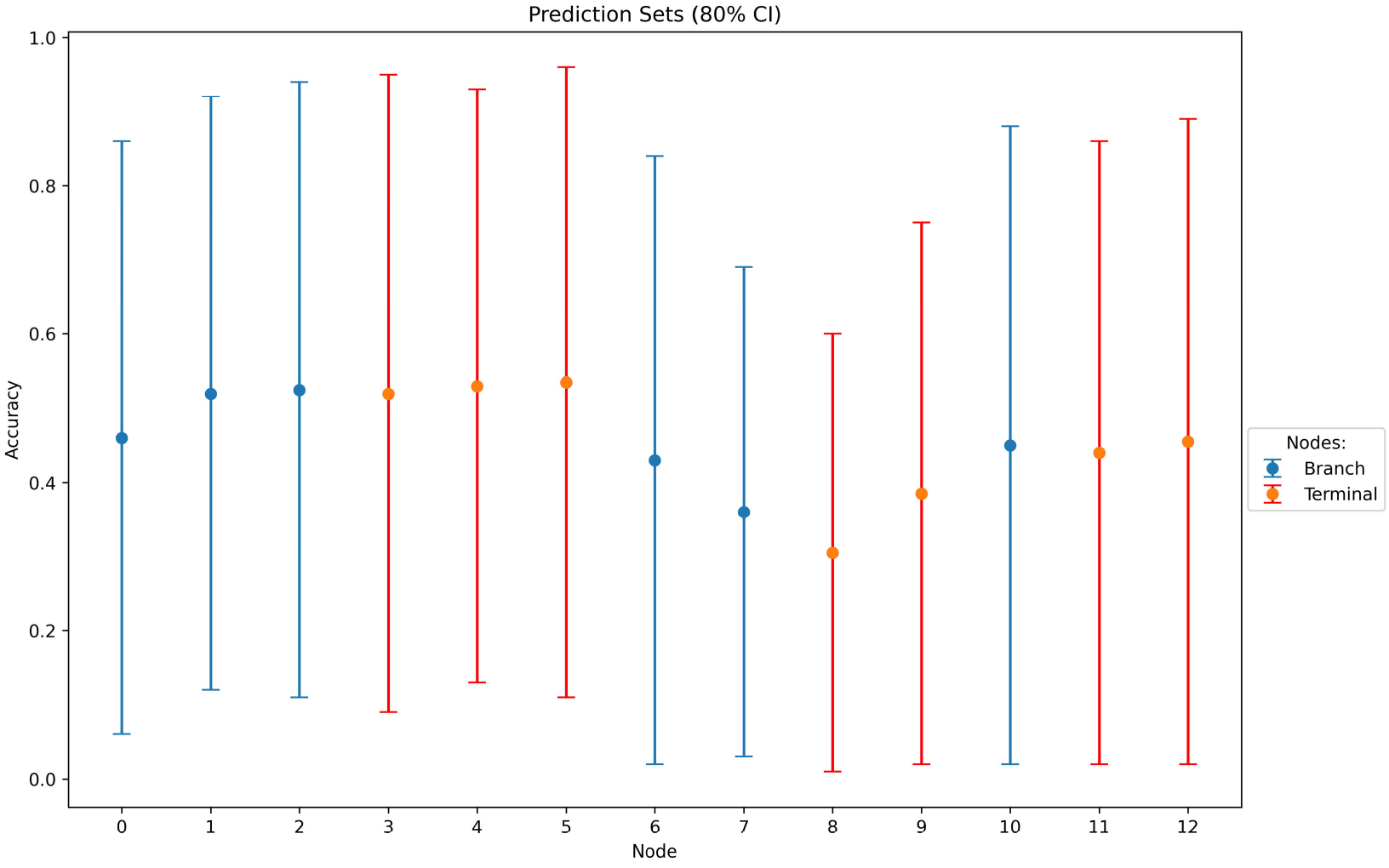}%
            \label{subfig:emory_biasresults_b}%
        }
        \hfill
        \subfloat[]{%
            \includegraphics[width=\twidth\linewidth]{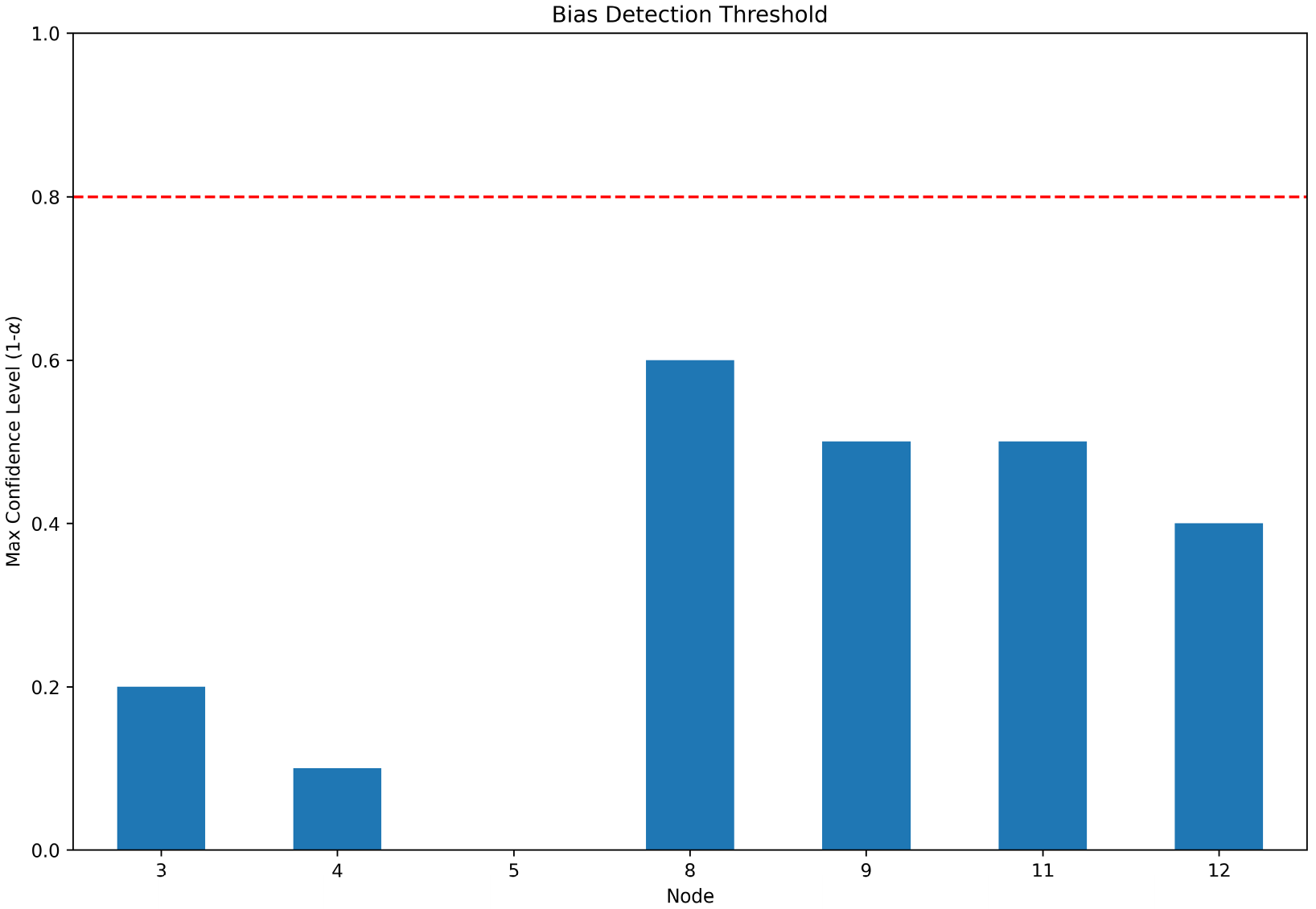}%
            \label{subfig:emory_biasresults_c}%
        }       
    \caption{Emory bias detection model results. \ref{subfig:emory_biasresults_a} displays the complete decision tree, \ref{subfig:emory_biasresults_b} shows the predicted confidence intervals $\hat{\mathcal{C}}_{j}$ for each branch (blue) and terminal (red) node at significance level $\alpha$. \ref{subfig:emory_biasresults_c} presents the maximum bias detection confidence level $1-\alpha^{*}_{j}$ for the $j^{\text{th}}$ terminal node.}
    \label{fig:emory_bias_results}
\end{figure}

The final results of our bias detection framework for the Emory University Hospital cohort are presented in Fig. \ref{fig:emory_bias_results}. Although Node 8 in Fig. \ref{subfig:emory_biasresults_a} represents the group of patients with the worst model performance, the confidence intervals shown in Fig. \ref{subfig:emory_biasresults_b} exhibit overlap across all terminal nodes. This overlap suggests \textit{that there is not enough evidence to indicate bias} in the model's performance for this cohort at significance level $\alpha^* = 0.20$. Furthermore, Fig. \ref{subfig:emory_biasresults_c} illustrates the optimized significance level $\alpha^*$ across all leaf nodes, indicating that bias would only be detected at $\alpha = 0.60$, corresponding to a confidence level of 0.40.


\section{Conclusion}
This paper introduces a novel approach to detecting and analyzing regions of algorithmic bias in medical-AI decision support systems. Our framework leverages the Classification and Regression Trees (CART) method, enhanced with conformal prediction intervals, to provide a robust mechanism for detecting and addressing potential biases in AI applications within the healthcare sector. We evaluated our technique through synthetic data experiments, demonstrating its capability to identify regions of bias, assuming such regions exist in the data. Furthermore, we extended our analysis to a real-world dataset by conducting an experiment using electronic health record (EHR) data obtained from Grady Memorial Hospital. The integration of conformal prediction intervals with the CART algorithm allows users to test a variety of confidence levels, thereby providing a flexible tool for determining the existence of algorithmic bias. By adjusting the confidence levels, users can explore the robustness of the bias detection across different thresholds, enhancing the reliability of the findings.

The increasing integration of machine learning and artificial intelligence in healthcare underscores the urgent need for tools, techniques, and procedures that ensure the fair and equitable use of these technologies. Our framework addresses this challenge by offering a practical solution for healthcare practitioners and AI developers to identify and mitigate algorithmic biases. This, in turn, promotes the development of medical ML/AI decision support systems that are both ethically sound and clinically effective.

\section*{Acknowledgement}
This work is partially supported by an NSF CAREER CCF-1650913, NSF DMS-2134037, CMMI-2015787, CMMI-2112533, DMS-1938106, DMS-1830210, NIGMS K23GM137182-03S1, Emory Hospital, and the Coca-Cola Foundation.

\bibliography{references}

\begin{thebibliography}{10}
\urlstyle{rm}
\expandafter\ifx\csname url\endcsname\relax
  \def\url#1{\texttt{#1}}\fi
\expandafter\ifx\csname urlprefix\endcsname\relax\def\urlprefix{URL }\fi
\expandafter\ifx\csname doiprefix\endcsname\relax\def\doiprefix{DOI: }\fi
\providecommand{\bibinfo}[2]{#2}
\providecommand{\eprint}[2][]{\url{#2}}

\bibitem{Ahmed2022ArtificialReview}
\bibinfo{author}{Ahmed, S.}, \bibinfo{author}{Alshater, M.~M.}, \bibinfo{author}{Ammari, A.~E.} \& \bibinfo{author}{Hammami, H.}
\newblock \bibinfo{journal}{\bibinfo{title}{{Artificial intelligence and machine learning in finance: A bibliometric review}}}.
\newblock {\emph{\JournalTitle{Research in International Business and Finance}}} \textbf{\bibinfo{volume}{61}}, \doiprefix\url{10.1016/j.ribaf.2022.101646} (\bibinfo{year}{2022}).

\bibitem{Dixon2020MachinePractice}
\bibinfo{author}{Dixon, M.~F.}, \bibinfo{author}{Halperin, I.} \& \bibinfo{author}{Bilokon, P.}
\newblock \emph{\bibinfo{title}{{Machine learning in finance: From theory to practice}}} (\bibinfo{publisher}{Springer}, \bibinfo{year}{2020}).

\bibitem{Kucak2018MachineTrends}
\bibinfo{author}{Ku{\v{c}}ak, D.}, \bibinfo{author}{Juri{\v{c}}i{\'{c}}, V.} \& \bibinfo{author}{Đambi{\'{c}}, G.}
\newblock \bibinfo{journal}{\bibinfo{title}{{Machine learning in education - A survey of current research trends}}}.
\newblock {\emph{\JournalTitle{Proceedings of the DAAAM International Scientific Conference}}} \textbf{\bibinfo{volume}{29}}, \bibinfo{pages}{059--067}, \doiprefix\url{10.2507/29th.daaam.proceedings.059} (\bibinfo{year}{2018}).

\bibitem{Luan2021AEducation}
\bibinfo{author}{Luan, H.} \& \bibinfo{author}{Tsai, C.~C.}
\newblock \bibinfo{journal}{\bibinfo{title}{{A Review of Using Machine Learning Approaches for Precision Education}}}.
\newblock {\emph{\JournalTitle{Educational Technology and Society}}} \textbf{\bibinfo{volume}{24}} (\bibinfo{year}{2021}).

\bibitem{Tiwari2023TheExperiences}
\bibinfo{author}{Tiwari, R.}
\newblock \bibinfo{journal}{\bibinfo{title}{{The integration of AI and machine learning in education and its potential to personalize and improve student learning experiences}}}.
\newblock {\emph{\JournalTitle{INTERANTIONAL JOURNAL OF SCIENTIFIC RESEARCH IN ENGINEERING AND MANAGEMENT}}} \textbf{\bibinfo{volume}{07}}, \doiprefix\url{10.55041/ijsrem17645} (\bibinfo{year}{2023}).

\bibitem{Broussard2023MachineSystem}
\bibinfo{author}{Broussard, M.}
\newblock \bibinfo{title}{{Machine Fairness and the Justice System}}.
\newblock In \emph{\bibinfo{booktitle}{More than a Glitch}}, \doiprefix\url{10.7551/mitpress/14234.003.0005} (\bibinfo{publisher}{MIT Press}, \bibinfo{year}{2023}).

\bibitem{Avila2020TheSystems}
\bibinfo{author}{{\'{A}}vila, F.}, \bibinfo{author}{Hannah-Moffat, K.} \& \bibinfo{author}{Maurutto, P. C. N. H. A.~.}
\newblock \bibinfo{title}{{The seductiveness of fairness: Is machine learning the answer? – Algorithmic fairness in criminal justice systems}}.
\newblock In \emph{\bibinfo{booktitle}{The algorithmic society: technology, power, and knowledge}}, \bibinfo{pages}{87--103} (\bibinfo{publisher}{Routledge}, \bibinfo{year}{2020}).

\bibitem{Chiao2019FairnessJustice}
\bibinfo{author}{Chiao, V.}
\newblock \bibinfo{title}{{Fairness, accountability and transparency: Notes on algorithmic decision-making in criminal justice}}, \doiprefix\url{10.1017/S1744552319000077} (\bibinfo{year}{2019}).

\bibitem{Obermeyer2019DissectingPopulations}
\bibinfo{author}{Obermeyer, Z.}, \bibinfo{author}{Powers, B.}, \bibinfo{author}{Vogeli, C.} \& \bibinfo{author}{Mullainathan, S.}
\newblock \bibinfo{journal}{\bibinfo{title}{{Dissecting racial bias in an algorithm used to manage the health of populations}}}.
\newblock {\emph{\JournalTitle{Science}}} \textbf{\bibinfo{volume}{366}}, \doiprefix\url{10.1126/science.aax2342} (\bibinfo{year}{2019}).

\bibitem{Pencina2020PredictionApplication}
\bibinfo{author}{Pencina, M.~J.}, \bibinfo{author}{Goldstein, B.~A.} \& \bibinfo{author}{D’Agostino, R.~B.}
\newblock \bibinfo{journal}{\bibinfo{title}{{Prediction Models — Development, Evaluation, and Clinical Application}}}.
\newblock {\emph{\JournalTitle{New England Journal of Medicine}}} \textbf{\bibinfo{volume}{382}}, \doiprefix\url{10.1056/nejmp2000589} (\bibinfo{year}{2020}).

\bibitem{Larson2016HowAlgorithm}
\bibinfo{author}{Larson, J.}, \bibinfo{author}{Mattu, S.}, \bibinfo{author}{Kirchner, L.} \& \bibinfo{author}{Angwin, J.}
\newblock \bibinfo{journal}{\bibinfo{title}{{How We Analyzed the COMPAS Recidivism Algorithm}}}.
\newblock {\emph{\JournalTitle{ProPublica}}}  (\bibinfo{year}{2016}).

\bibitem{Larrazabal2020GenderDiagnosis}
\bibinfo{author}{Larrazabal, A.~J.}, \bibinfo{author}{Nieto, N.}, \bibinfo{author}{Peterson, V.}, \bibinfo{author}{Milone, D.~H.} \& \bibinfo{author}{Ferrante, E.}
\newblock \bibinfo{journal}{\bibinfo{title}{{Gender imbalance in medical imaging datasets produces biased classifiers for computer-aided diagnosis}}}.
\newblock {\emph{\JournalTitle{Proceedings of the National Academy of Sciences of the United States of America}}} \textbf{\bibinfo{volume}{117}}, \doiprefix\url{10.1073/pnas.1919012117} (\bibinfo{year}{2020}).

\bibitem{Gianfrancesco2018PotentialData}
\bibinfo{author}{Gianfrancesco, M.~A.}, \bibinfo{author}{Tamang, S.}, \bibinfo{author}{Yazdany, J.} \& \bibinfo{author}{Schmajuk, G.}
\newblock \bibinfo{title}{{Potential Biases in Machine Learning Algorithms Using Electronic Health Record Data}}, \doiprefix\url{10.1001/jamainternmed.2018.3763} (\bibinfo{year}{2018}).

\bibitem{Dwork2012FairnessAwareness}
\bibinfo{author}{Dwork, C.}, \bibinfo{author}{Hardt, M.}, \bibinfo{author}{Pitassi, T.}, \bibinfo{author}{Reingold, O.} \& \bibinfo{author}{Zemel, R.}
\newblock \bibinfo{title}{{Fairness through awareness}}.
\newblock In \emph{\bibinfo{booktitle}{ITCS 2012 - Innovations in Theoretical Computer Science Conference}}, \doiprefix\url{10.1145/2090236.2090255} (\bibinfo{year}{2012}).

\bibitem{Kusner2017CounterfactualFairness}
\bibinfo{author}{Kusner, M.}, \bibinfo{author}{Loftus, J.}, \bibinfo{author}{Russell, C.} \& \bibinfo{author}{Silva, R.}
\newblock \bibinfo{title}{{Counterfactual fairness}}.
\newblock In \emph{\bibinfo{booktitle}{Advances in Neural Information Processing Systems}}, vol. \bibinfo{volume}{2017-December} (\bibinfo{year}{2017}).

\bibitem{Narayanan2018Tutorial:Politics}
\bibinfo{author}{Narayanan, A.}
\newblock \bibinfo{journal}{\bibinfo{title}{{Tutorial: 21 Fairness Definitions and their Politics}}}.
\newblock {\emph{\JournalTitle{Conference on Fairiness, Accountability, and Transparency}}}  (\bibinfo{year}{2018}).

\bibitem{Castelnovo2022ALandscape}
\bibinfo{author}{Castelnovo, A.} \emph{et~al.}
\newblock \bibinfo{journal}{\bibinfo{title}{{A clarification of the nuances in the fairness metrics landscape}}}.
\newblock {\emph{\JournalTitle{Scientific Reports}}} \textbf{\bibinfo{volume}{12}}, \doiprefix\url{10.1038/s41598-022-07939-1} (\bibinfo{year}{2022}).

\bibitem{Hardt2016EqualityLearning}
\bibinfo{author}{Hardt, M.}, \bibinfo{author}{Price, E.} \& \bibinfo{author}{Srebro, N.}
\newblock \bibinfo{title}{{Equality of opportunity in supervised learning}}.
\newblock In \emph{\bibinfo{booktitle}{Advances in Neural Information Processing Systems}} (\bibinfo{year}{2016}).

\bibitem{Chouldechova2017FairInstruments}
\bibinfo{author}{Chouldechova, A.}
\newblock \bibinfo{journal}{\bibinfo{title}{{Fair Prediction with Disparate Impact: A Study of Bias in Recidivism Prediction Instruments}}}.
\newblock {\emph{\JournalTitle{Big Data}}} \textbf{\bibinfo{volume}{5}}, \doiprefix\url{10.1089/big.2016.0047} (\bibinfo{year}{2017}).

\bibitem{Feldman2015CertifyingImpact}
\bibinfo{author}{Feldman, M.}, \bibinfo{author}{Friedler, S.~A.}, \bibinfo{author}{Moeller, J.}, \bibinfo{author}{Scheidegger, C.} \& \bibinfo{author}{Venkatasubramanian, S.}
\newblock \bibinfo{title}{{Certifying and removing disparate impact}}.
\newblock In \emph{\bibinfo{booktitle}{Proceedings of the ACM SIGKDD International Conference on Knowledge Discovery and Data Mining}}, vol. \bibinfo{volume}{2015-August}, \doiprefix\url{10.1145/2783258.2783311} (\bibinfo{year}{2015}).

\bibitem{Dwork2019FairnessComposition}
\bibinfo{author}{Dwork, C.} \& \bibinfo{author}{Ilvento, C.}
\newblock \bibinfo{title}{{Fairness under composition}}.
\newblock In \emph{\bibinfo{booktitle}{Leibniz International Proceedings in Informatics, LIPIcs}}, vol. \bibinfo{volume}{124}, \doiprefix\url{10.4230/LIPIcs.ITCS.2019.33} (\bibinfo{year}{2019}).

\bibitem{Crenshaw2018Demarginalizing1989}
\bibinfo{author}{Crenshaw, K.}
\newblock \bibinfo{journal}{\bibinfo{title}{{Demarginalizing the intersection of race and sex: A black feminist critique of antidiscrimination doctrine, feminist theory, and antiracist politics [1989]}}}.
\newblock {\emph{\JournalTitle{Feminist Legal Theory: Readings in Law and Gender}}} \bibinfo{pages}{139--167}, \doiprefix\url{10.4324/9780429500480} (\bibinfo{year}{2018}).

\bibitem{Gohar2023AChallenges}
\bibinfo{author}{Gohar, U.} \& \bibinfo{author}{Cheng, L.}
\newblock \bibinfo{title}{{A Survey on Intersectional Fairness in Machine Learning: Notions, Mitigation, and Challenges}}.
\newblock In \emph{\bibinfo{booktitle}{IJCAI International Joint Conference on Artificial Intelligence}}, vol. \bibinfo{volume}{2023-August}, \doiprefix\url{10.24963/ijcai.2023/742} (\bibinfo{year}{2023}).

\bibitem{Kearns2018PreventingFairness}
\bibinfo{author}{Kearns, M.}, \bibinfo{author}{Neel, S.}, \bibinfo{author}{Roth, A.} \& \bibinfo{author}{Wu, Z.~S.}
\newblock \bibinfo{title}{{Preventing fairness gerrymandering: Auditing and learning for subgroup fairness}}.
\newblock In \emph{\bibinfo{booktitle}{35th International Conference on Machine Learning, ICML 2018}}, vol.~\bibinfo{volume}{6} (\bibinfo{year}{2018}).

\bibitem{Hebert-Johnson2018Multicalibration:Masses}
\bibinfo{author}{Hebert-Johnson, U.}, \bibinfo{author}{Kim, M.~P.}, \bibinfo{author}{Reingold, O.} \& \bibinfo{author}{Rothblum, G.~N.}
\newblock \bibinfo{title}{{Multicalibration: Calibration for the (computationally-identifiable) masses}}.
\newblock In \emph{\bibinfo{booktitle}{35th International Conference on Machine Learning, ICML 2018}}, vol.~\bibinfo{volume}{5} (\bibinfo{year}{2018}).

\bibitem{Pastor2021IdentifyingClassification}
\bibinfo{author}{Pastor, E.}, \bibinfo{author}{de~Alfaro, L.} \& \bibinfo{author}{Baralis, E.}
\newblock \bibinfo{journal}{\bibinfo{title}{{Identifying Biased Subgroups in Ranking and Classification}}}.
\newblock {\emph{\JournalTitle{In Responsible AI @ KDD 2021 Workshop}}}  (\bibinfo{year}{2021}).

\bibitem{Chen2004FailureTrees}
\bibinfo{author}{Chen, M.}, \bibinfo{author}{Zheng, A.~X.}, \bibinfo{author}{Lloyd, J.}, \bibinfo{author}{Jordan, M.~I.} \& \bibinfo{author}{Brewer, E.}
\newblock \bibinfo{title}{{Failure diagnosis using decision trees}}.
\newblock In \emph{\bibinfo{booktitle}{Proceedings - International Conference on Autonomic Computing}}, \doiprefix\url{10.1109/ICAC.2004.1301345} (\bibinfo{year}{2004}).

\bibitem{Singla2021UnderstandingExtraction}
\bibinfo{author}{Singla, S.}, \bibinfo{author}{Nushi, B.}, \bibinfo{author}{Shah, S.}, \bibinfo{author}{Kamar, E.} \& \bibinfo{author}{Horvitz, E.}
\newblock \bibinfo{title}{{Understanding failures of deep networks via robust feature extraction}}.
\newblock In \emph{\bibinfo{booktitle}{Proceedings of the IEEE Computer Society Conference on Computer Vision and Pattern Recognition}}, \doiprefix\url{10.1109/CVPR46437.2021.01266} (\bibinfo{year}{2021}).

\bibitem{Nushi2018TowardsFailure}
\bibinfo{author}{Nushi, B.}, \bibinfo{author}{Kamar, E.} \& \bibinfo{author}{Horvitz, E.}
\newblock \bibinfo{title}{{Towards Accountable AI: Hybrid Human-Machine Analyses for Characterizing System Failure}}.
\newblock In \emph{\bibinfo{booktitle}{Proceedings of the 6th AAAI Conference on Human Computation and Crowdsourcing, HCOMP 2018}}, \doiprefix\url{10.1609/hcomp.v6i1.13337} (\bibinfo{year}{2018}).

\bibitem{Breiman2017ClassificationTrees}
\bibinfo{author}{Breiman, L.}, \bibinfo{author}{Friedman, J.~H.}, \bibinfo{author}{Olshen, R.~A.} \& \bibinfo{author}{Stone, C.~J.}
\newblock \emph{\bibinfo{title}{{Classification and regression trees}}} (\bibinfo{publisher}{Chapman {\&}Hall/CRC}, \bibinfo{year}{2017}).

\bibitem{Chipman1998BayesianSearch}
\bibinfo{author}{Chipman, H.~A.}, \bibinfo{author}{George, E.~I.} \& \bibinfo{author}{McCulloch, R.~E.}
\newblock \bibinfo{journal}{\bibinfo{title}{{Bayesian CART model search}}}.
\newblock {\emph{\JournalTitle{Journal of the American Statistical Association}}} \textbf{\bibinfo{volume}{93}}, \doiprefix\url{10.1080/01621459.1998.10473750} (\bibinfo{year}{1998}).

\bibitem{Singer2016Thesepsis-3}
\bibinfo{author}{Singer, M.} \emph{et~al.}
\newblock \bibinfo{title}{{The third international consensus definitions for sepsis and septic shock (sepsis-3)}}, \doiprefix\url{10.1001/jama.2016.0287} (\bibinfo{year}{2016}).

\bibitem{Jones2009ThePresentation}
\bibinfo{author}{Jones, A.~E.}, \bibinfo{author}{Trzeciak, S.} \& \bibinfo{author}{Kline, J.~A.}
\newblock \bibinfo{journal}{\bibinfo{title}{{The Sequential Organ Failure Assessment score for predicting outcome in patients with severe sepsis and evidence of hypoperfusion at the time of emergency department presentation}}}.
\newblock {\emph{\JournalTitle{Critical Care Medicine}}} \textbf{\bibinfo{volume}{37}}, \doiprefix\url{10.1097/CCM.0b013e31819def97} (\bibinfo{year}{2009}).

\bibitem{Yang2019EarlyOptimization}
\bibinfo{author}{Yang, M.} \emph{et~al.}
\newblock \bibinfo{title}{{Early Prediction of Sepsis Using Multi-Feature Fusion Based XGBoost Learning and Bayesian Optimization}}.
\newblock In \emph{\bibinfo{booktitle}{2019 Computing in Cardiology Conference (CinC)}}, vol.~\bibinfo{volume}{45}, \doiprefix\url{10.22489/cinc.2019.020} (\bibinfo{year}{2019}).

\bibitem{Groenwold2020InformativeKnowing}
\bibinfo{author}{Groenwold, R. H.~H.}
\newblock \bibinfo{journal}{\bibinfo{title}{{Informative missingness in electronic health record systems: the curse of knowing}}}.
\newblock {\emph{\JournalTitle{Diagnostic and Prognostic Research}}} \textbf{\bibinfo{volume}{4}}, \doiprefix\url{10.1186/s41512-020-00077-0} (\bibinfo{year}{2020}).

\bibitem{Vincent1996TheDysfunction/failure}
\bibinfo{author}{Vincent, J.~L.} \emph{et~al.}
\newblock \bibinfo{journal}{\bibinfo{title}{{The SOFA (Sepsis-related Organ Failure Assessment) score to describe organ dysfunction/failure}}}.
\newblock {\emph{\JournalTitle{Intensive Care Medicine}}} \textbf{\bibinfo{volume}{22}}, \doiprefix\url{10.1007/BF01709751} (\bibinfo{year}{1996}).

\bibitem{Smith2013TheDeath}
\bibinfo{author}{Smith, G.~B.}, \bibinfo{author}{Prytherch, D.~R.}, \bibinfo{author}{Meredith, P.}, \bibinfo{author}{Schmidt, P.~E.} \& \bibinfo{author}{Featherstone, P.~I.}
\newblock \bibinfo{journal}{\bibinfo{title}{{The ability of the National Early Warning Score (NEWS) to discriminate patients at risk of early cardiac arrest, unanticipated intensive care unit admission, and death}}}.
\newblock {\emph{\JournalTitle{Resuscitation}}} \textbf{\bibinfo{volume}{84}}, \doiprefix\url{10.1016/j.resuscitation.2012.12.016} (\bibinfo{year}{2013}).

\bibitem{Machado2016GettingSettings}
\bibinfo{author}{Machado, F.~R.} \emph{et~al.}
\newblock \bibinfo{title}{{Getting a consensus: Advantages and disadvantages of Sepsis 3 in the context of middle-income settings}}, \doiprefix\url{10.5935/0103-507X.20160068} (\bibinfo{year}{2016}).

\bibitem{Chen2016XGBoost:System}
\bibinfo{author}{Chen, T.} \& \bibinfo{author}{Guestrin, C.}
\newblock \bibinfo{title}{{XGBoost: A scalable tree boosting system}}.
\newblock In \emph{\bibinfo{booktitle}{Proceedings of the ACM SIGKDD International Conference on Knowledge Discovery and Data Mining}}, vol. \bibinfo{volume}{13-17-August-2016}, \doiprefix\url{10.1145/2939672.2939785} (\bibinfo{year}{2016}).

\bibitem{Bergstra2011AlgorithmsOptimization}
\bibinfo{author}{Bergstra, J.}, \bibinfo{author}{Bardenet, R.}, \bibinfo{author}{Bengio, Y.} \& \bibinfo{author}{K{\'{e}}gl, B.}
\newblock \bibinfo{title}{{Algorithms for hyper-parameter optimization}}.
\newblock In \emph{\bibinfo{booktitle}{Advances in Neural Information Processing Systems 24: 25th Annual Conference on Neural Information Processing Systems 2011, NIPS 2011}} (\bibinfo{year}{2011}).

\end{thebibliography}
\appendix

\section{Data pre-processing} \label{appenx:data_pre_proc}
These datasets include a diverse range of continuous physiological measurements, vital signs, laboratory results, and medical treatment information for each encounter. Data also incorporated demographic information from the patient, including age, sex, race, zip code, and insurance status, which we utilize in later stages of the study. We perform feature reduction by removing physiological features missing more than 75\% of their records. This resulted in 39 continuous patient features remaining for analysis as denoted by Table \ref{table:features}. In addition, we included two administrative identifiers: procedure and ventilation status.
\begin{table}[h!]
    \caption{Patient physiologic features selected for analysis}
    \centering
    \resizebox{\columnwidth}{!}{
    \begin{tabular}{c|c| c} \hline
    Vitals (8) &  \multicolumn{2}{c}{Labs (31)} \\ \hline
    Best Mean Arterial Pressure (MAP)  & Alanine Aminotransfer & Hematocrit  \\
    Heart Rate (HR) & Albumin & Hemoglobin \\
    Oxygen Saturation (SpO2) & Alkaline Phosphatase & Magnesium\\
    Respiratory Rate & Anion Gap & Partial Pressure of Carbon Dioxide (PaCO2) \\
    Temperature & Aspartate Aminotransferase (AST) & Partial Pressure of Oxygen (PaO2) \\
    Systolic Blood Pressure (Cuff) & Base Excess &  Partial Pressure of Oxygen/Fraction of Blood Oxygen Saturation (p/F Ratio)\\
    Diastolic Blood Pressure (Cuff) & Bicarb (HCO3) & pH\\
    Mean Arterial Pressure (Cuff) & Bilirubin Total & Phosphorus \\
    & Blood Urea Nitrogen (BUN) & Platelets\\
    & Calcium & Potassium\\
    & Chloride & Protein\\
    & Creatinine & Sodium\\
    & Daily Weight kg & White Blood Cell Count\\
    & FiO2 &  SOFA Score Total\\
    & Glasgow Coma Score (total) &  SIRS Score Total\\
    & Glucose &  \\
    
    \hline
    
    \end{tabular}
        }
    \label{table:features}
\end{table}

We impute missing data through a forward-filling approach. When a feature $x$ has a previously recorded value, $v$, at time step $t_p < t$, we set $x_{v}^{(t)} = x_v^{t_p}$ to forward-fill the missing value of $v$ at time step $t$. If no prior recorded value exists, the missing value remains unprocessed. Lastly, to mitigate data leakage, we remove sepsis patient data following their first retrospectively identified sepsis hour.

\subsection{Feature engineering}
Following our initial data pre-processing, which resulted in 41 selected physiological patient features, we further develop three categories of variables in this section. These include 72 variables for indicating the informativeness of missing features, 89 time-series based features, and eight clinically relevant features for assessing sepsis. The final dataset, following all feature engineering steps, resulted in a total of 210 features.

\subsubsection*{Feature informative missingness}
The presence of missing data, a common occurrence in routinely collected health information, can provide significant insights, as the nature of the missing data itself can be informative \cite{Groenwold2020InformativeKnowing}. The collection times for clinical laboratory and treatment information fluctuate among individuals and may vary throughout their treatment period, resulting in a significant number of missing entries in the physiological data, including instances where entire features are absent. This phenomenon of missing data, particularly prevalent in ICU settings, is not without pattern as it often reflects the clinical judgments made regarding a patient's critical condition.  We introduce two missing data indicator sequences for 36 specific variables, which include all lab values, ventilation status, systolic blood pressure, diastolic blood pressure, and mean arterial pressure, with the aim to harness the latent predictive value embedded within these missing data points. The {\it Measurement Frequency} (f1) sequence counts the number of measurements taken for a variable before the current time. The {\it Measurement Time Interval} (f2) sequence records the time interval from the most recent measurement to the current time. A value of $-1$ is assigned when there is no prior recorded measurement.

Table \ref{table:fim} illustrates the application of two missing data indicator sequences through an example of an eight-hour time series for temperature measurements. The first row displays the temperature readings over time. The second row shows the measurement frequency sequence, indicating the cumulative number of temperature measurements taken up to each point in time. The final row presents the measurement time interval sequence, highlighting the time elapsed since the last temperature measurement, with a notation of -1 when there is no previous measurement to reference.
\begin{table}[h!]
    \caption{Example of feature informative missingness sequences}
    \centering
    \resizebox{.5\columnwidth}{!}{
    \begin{tabular}{c|c|c|c|c|c|c|c|c} \hline
     & nan & 98.0 & 98.1 & nan & nan & 98.2 & nan &  97.4 \\ \hline 
   f1 score & 0 & 1 & 2 & 2 & 2 & 3 & 3 & 4 \\ \hline
    f2 score & $-1$ & 0 & 0 & 1 & 2 & 0 & 1 & 0\\
    
    \hline
    
    \end{tabular}
        }
    \label{table:fim}
\end{table}
\subsubsection*{Clinical empiric features}

Historically, rule-based severity scoring systems for diseases like the Sequential Organ Failure Assessment (SOFA) \cite{Vincent1996TheDysfunction/failure}, quick-SOFA (qSOFA) \cite{Singer2016Thesepsis-3}, and the National Early Warning System (NEWS) \cite{Smith2013TheDeath} have been used to define sepsis in clinical settings. However, these systems may not satisfy the critical need for timely detection of sepsis to initiate effective treatment \cite{Machado2016GettingSettings}. We highlight the importance of several measurements to quantify abnormalities according to some scoring system. The qSOFA score is identified as ``1'' with Systolic BP (SBP) $\leq$ 100 mm Hg and Respiration rate (Resp) $\geq$ 22/min, otherwise ``0''. The measurements of platelets, bilirubin, mean arterial pressure (MAP), and creatinine are scored respectively under the rules of SOFA score, while heart rate, temperature, and respiration rate are scored on the basis of the NEWS score.

\subsubsection*{Time series features}
To capture the dynamic changes in patients' data records, we calculate two types of time-series features as follows.
\begin{itemize}
    \item {\it Differential features}: These are derived by computing the difference between the current value and the previous measurement of a given feature. This calculation highlights the immediate changes in patient conditions.
    
    \item {\it Sliding-window-based statistical features}: For this analysis, we focus on eight vital sign measurements: Best Mean Arterial Pressure (MAP), Heart Rate (HR), Oxygen Saturation (SpO2), Respiratory Rate, Temperature, Diastolic Blood Pressure (DBP), Systolic Blood Pressure (SBP), and Mean Arterial Pressure (MAP). We employ a fixed-length rolling six-hour sliding window to segment each record. This fixed rolling window increments in one-hour steps. In instances where the window is less than six hours, the sliding window includes all available data. Finally, we calculate key statistical features for each window, including maximum, minimum, mean, median, standard deviation, and differential standard deviation for each of the selected measurements.
\end{itemize}
\subsubsection*{Sepsis label lead time}
This study aims to develop a prognostic model that can accurately predict the onset of sepsis up to six hours before it happens. To highlight the significance of identifying sepsis at an early stage, we have introduced a six-hour lead time on the sepsis indicator variable. This adjustment enables the model to specifically focus on and recognize probable sepsis cases before they completely develop, thereby improving the model's ability to forecast outcomes in clinical settings.

\section{XGBoost Model} \label{appendix:xgb_model}

The sepsis prediction model developed for this analysis was centered on the implementation of XGBoost \cite{Chen2016XGBoost:System}, a robust tree-based gradient boosting algorithm known for its high computational efficiency and exceptional performance in managing complex and large datasets. We constructed this model using the Bayesian optimization technique with a Tree-structured Parzen Estimator (TPE) \cite{Bergstra2011AlgorithmsOptimization} approach. We applied this method to optimize hyperparameters, which helped establish the learning process, complexity, and generalization capability of the model. Hyperparameters included but were not limited to, the following: max depth, learning rate, and alpha and lambda regularization terms. 

The Bayesian optimization technique involved a series of 20 evaluations. In each iteration, we tune the hyperparameters with the aim of maximizing the accuracy of the prediction model. The final model is an ensemble based on the average five-fold cross-validation performance measured across this accuracy optimized loss function.

\subsection{Training, validation, and test sets}
In crafting our machine learning model, we incorporated a nuanced approach that integrates stratified cross-validation, temporal partitioning of data, and ensemble techniques to address the inherent challenges of predicting sepsis through the use of temporal dataset. This framework is specifically designed to evaluate models on future, unobserved data, thus closely simulating real-world clinical forecasting scenarios and enhancing the model's external validity. Our stratification strategy ensures that each subset for training and validation is a representative sample of the entire dataset by addressing class imbalance across folds. We incorporate an ensemble methodology to leverage the collective insights from multiple models, with the aim to reduce variability and enhance the reliability across predictions.

To construct our training, validation, and testing datasets we initially divided the dataset temporally, creating two groups: one with patients admitted to the ICU prior to 2019, designated for training and validation purposes, and the other comprised of patients from 2019 onwards for testing. Within the pre-2019 dataset, we performed stratified five-fold cross-validation to further partition the data into five exhaustive and mutually exclusive subsets. We execute this stratification with respect to the sepsis label to guarantee that each fold contains a proportional distribution of cases, both septic and non-septic. 
 
\begin{figure}[h!]
    \centering
    \includegraphics[width=\textwidth]{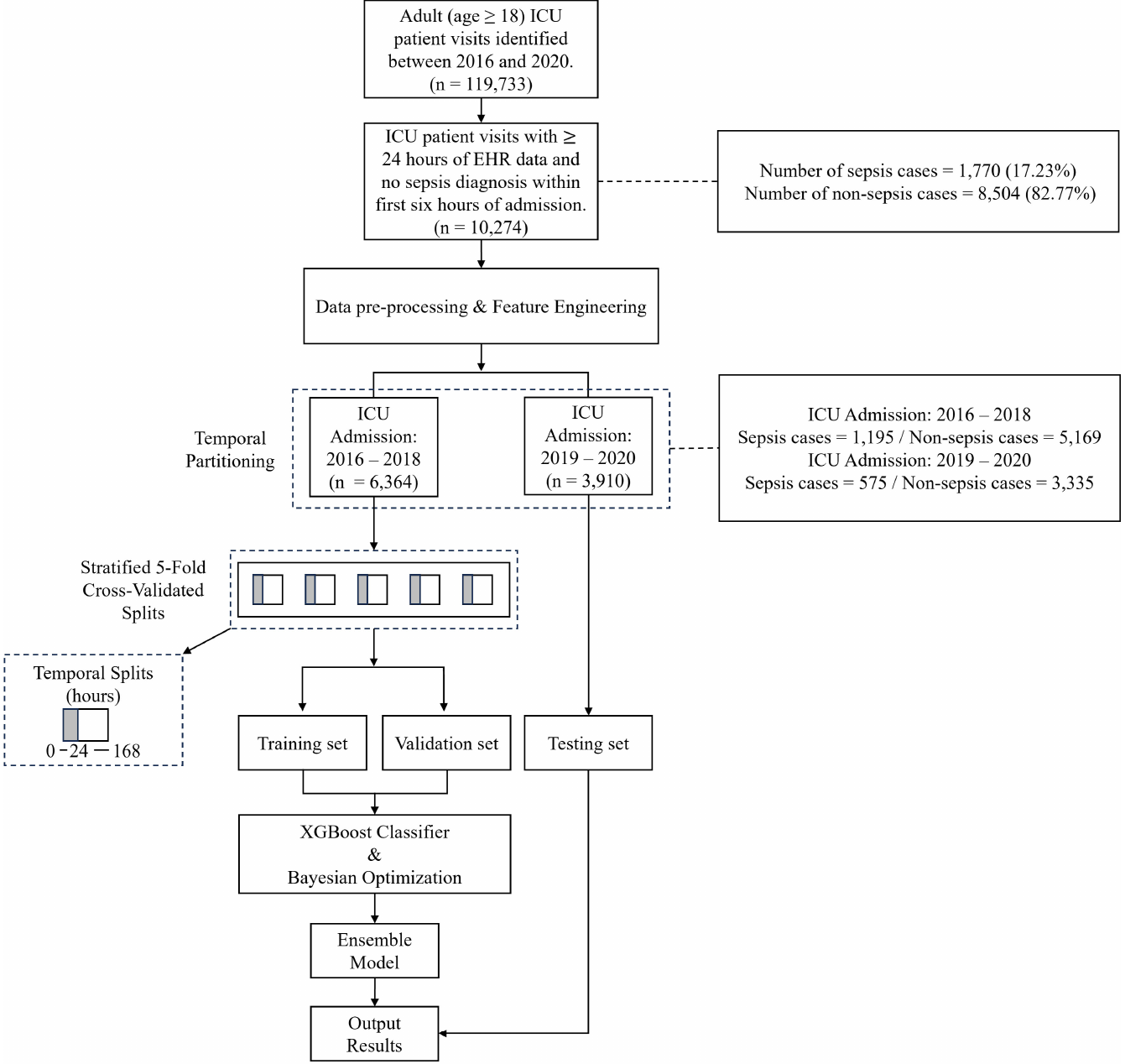}
    \caption{Illustration of the data pre-processing and model development procedure of the Grady sepsis prediction model.}
    \label{fig:model_development}
\end{figure}

 Within each of these five stratified folds we include all relevant continuous physiological data for each patient, reflecting the previously mentioned comprehensive feature engineering process that was undertaken. We further temporally partition this data, allocating the initial 24 hours of records following a patient's admission to the ICU to the training set, and the subsequent records, up to the 168th hour, to the validation set. This 168-hour cap is strategically selected to reduce the potential impacts of data bias that might arise from complications affecting a patient's health status beyond the initial week of their ICU stay. To address the imbalance between sepsis and non-sepsis hours, we also undertake a down sampling of the non-sepsis instances within each fold. Each fold thus generates a model trained on its designated training data subset and validated on its respective validation set. Collectively, these models form an ensemble, capitalizing on the variability and strengths of each model trained and validated on slightly different data segments. Fig. \ref{fig:model_development} depicts the complete data pre-processing and model development pipeline using the Grady dataset.

\subsection{Model results}
Table \ref{fig:ens_model_results} provides a comparative summary of the performance of individual XGBoost models and the ensemble model across both cohorts—Grady Memorial Hospital and Emory University Hospital. The table presents a horizontal comparison, reporting the accuracy and area under the curve (AUC) for each model across each cross-validation fold.

\begin{figure}[!htbp]
    \def\twidth{0.25}
    \centering
        \subfloat[]{%
            \includegraphics[width=\twidth\linewidth]{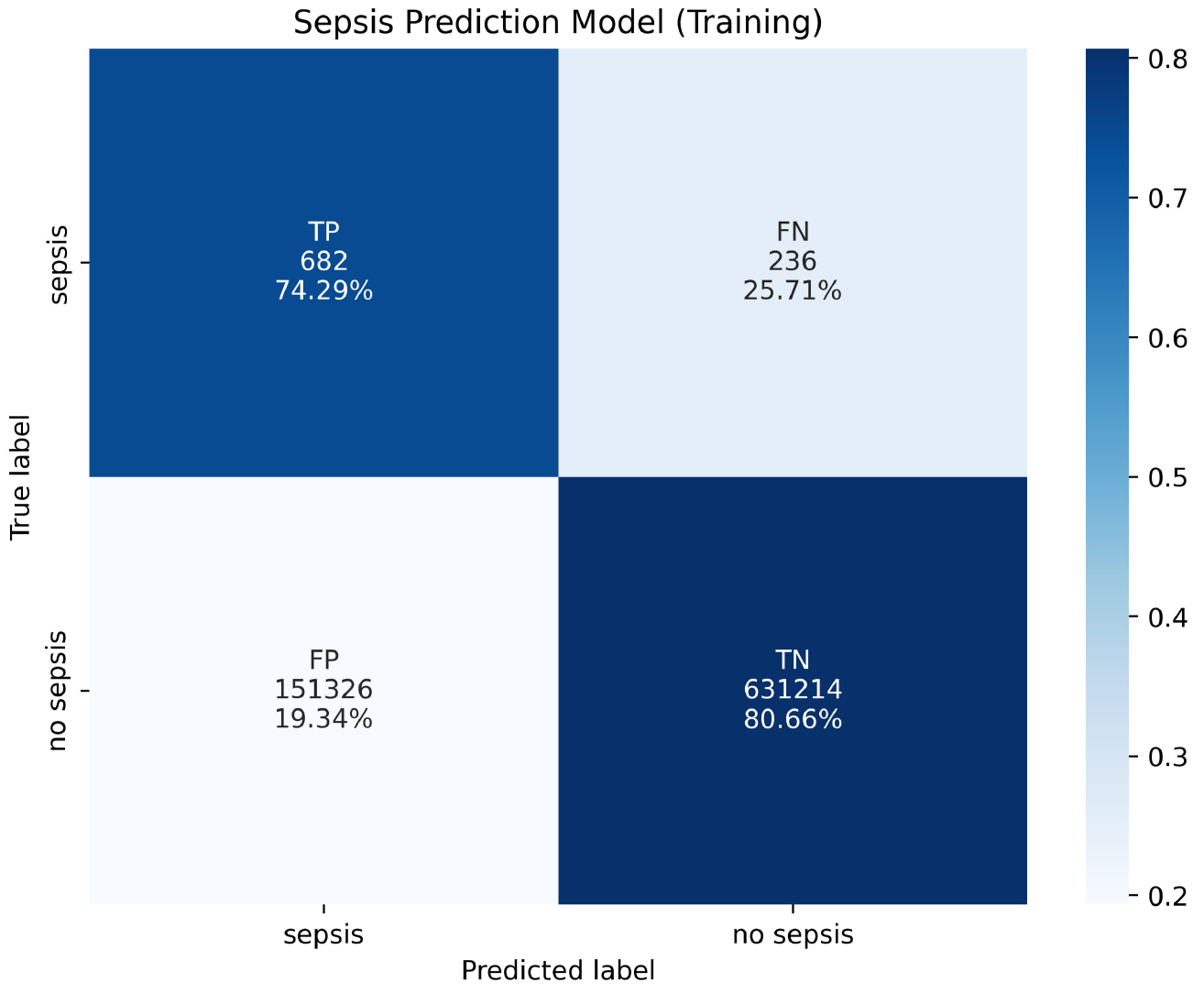}%
            \label{subfig:sepsis_pred_a}%
        }
        \subfloat[]{%
            \includegraphics[width=\twidth\linewidth]{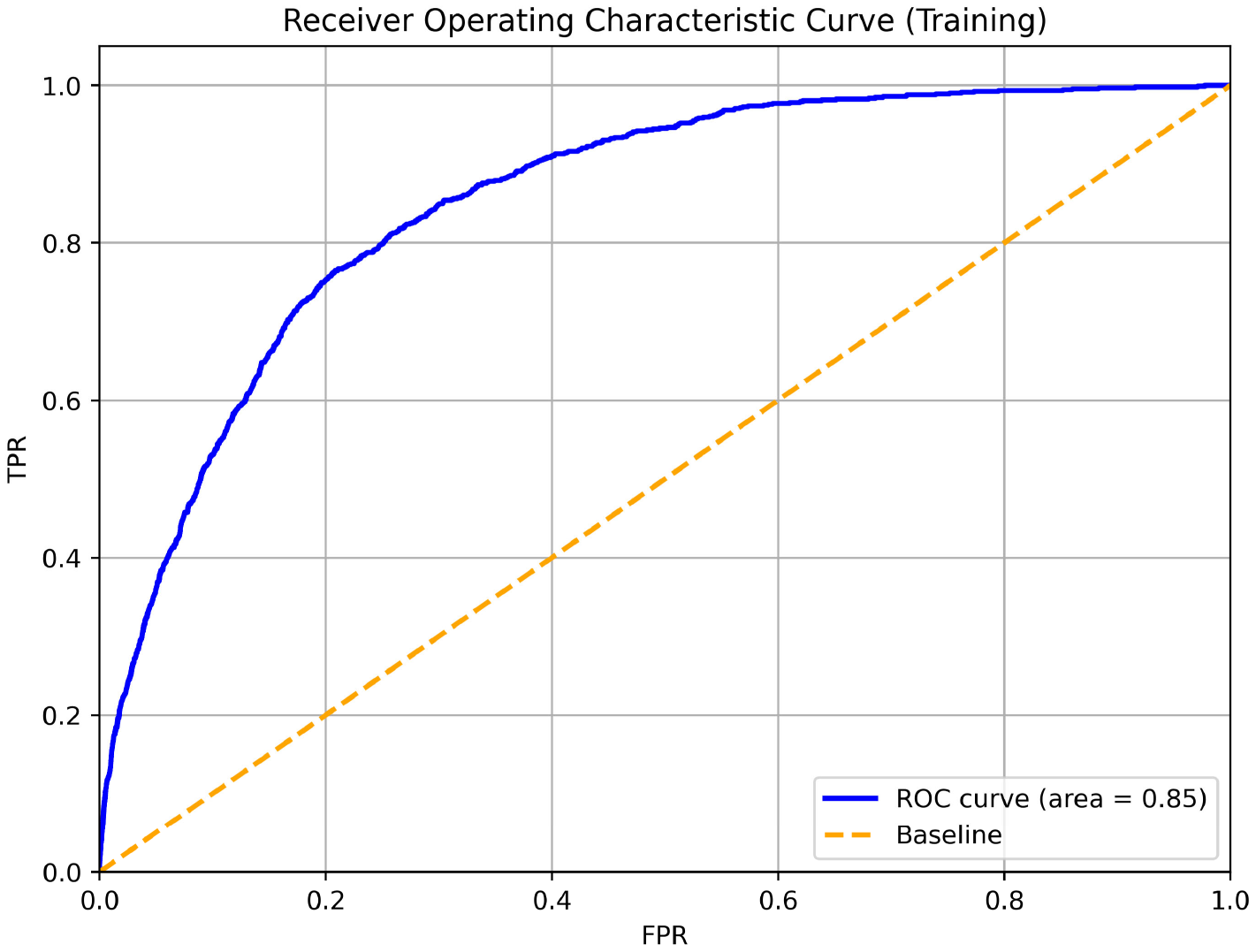}%
            \label{subfig:sepsis_pred_b}%
        }
        \subfloat[]{%
            \includegraphics[width=.25\linewidth]{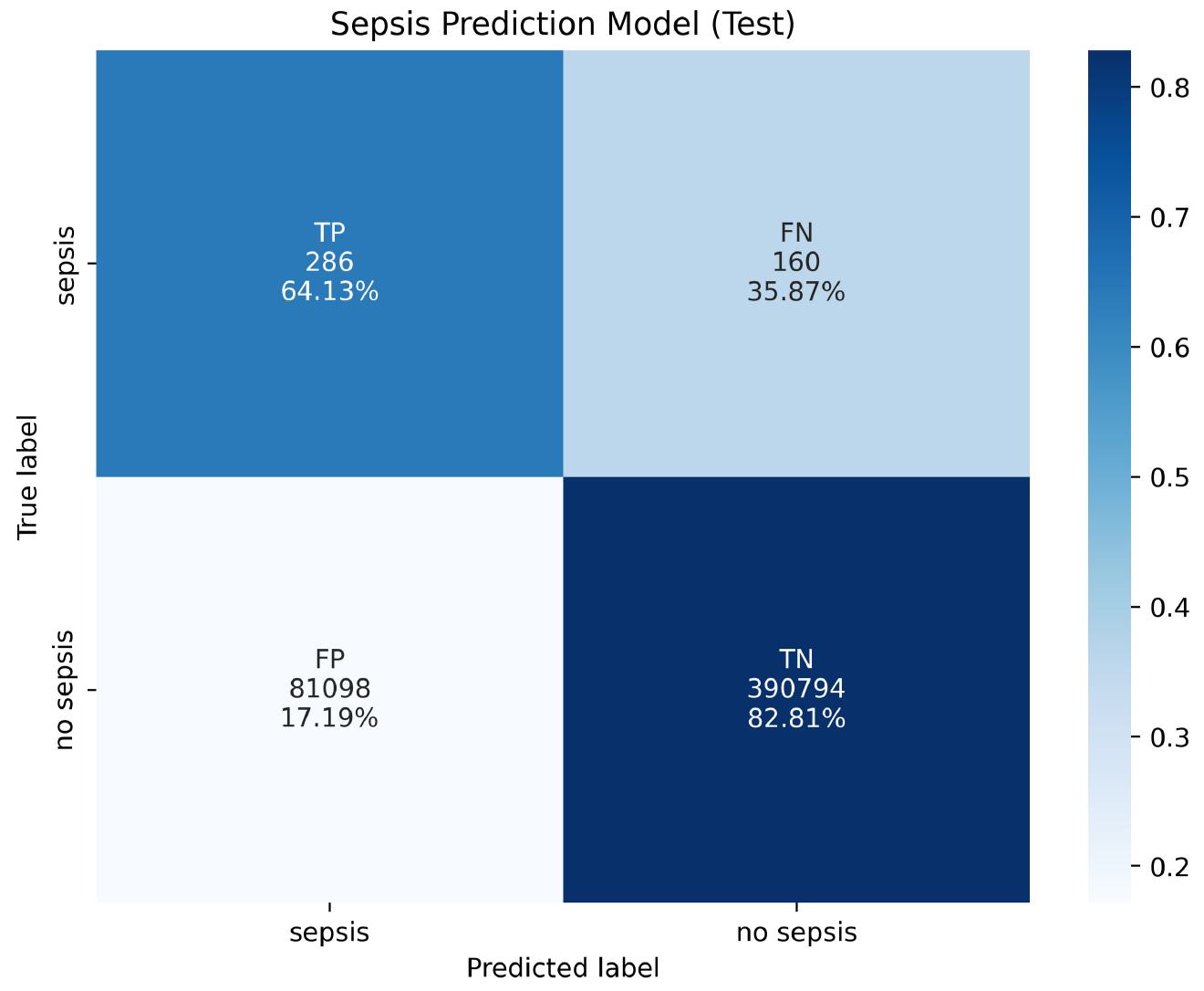}%
            \label{subfig:sepsis_pred_c}%
        }
        \subfloat[]{%
            \includegraphics[width=.25\linewidth]{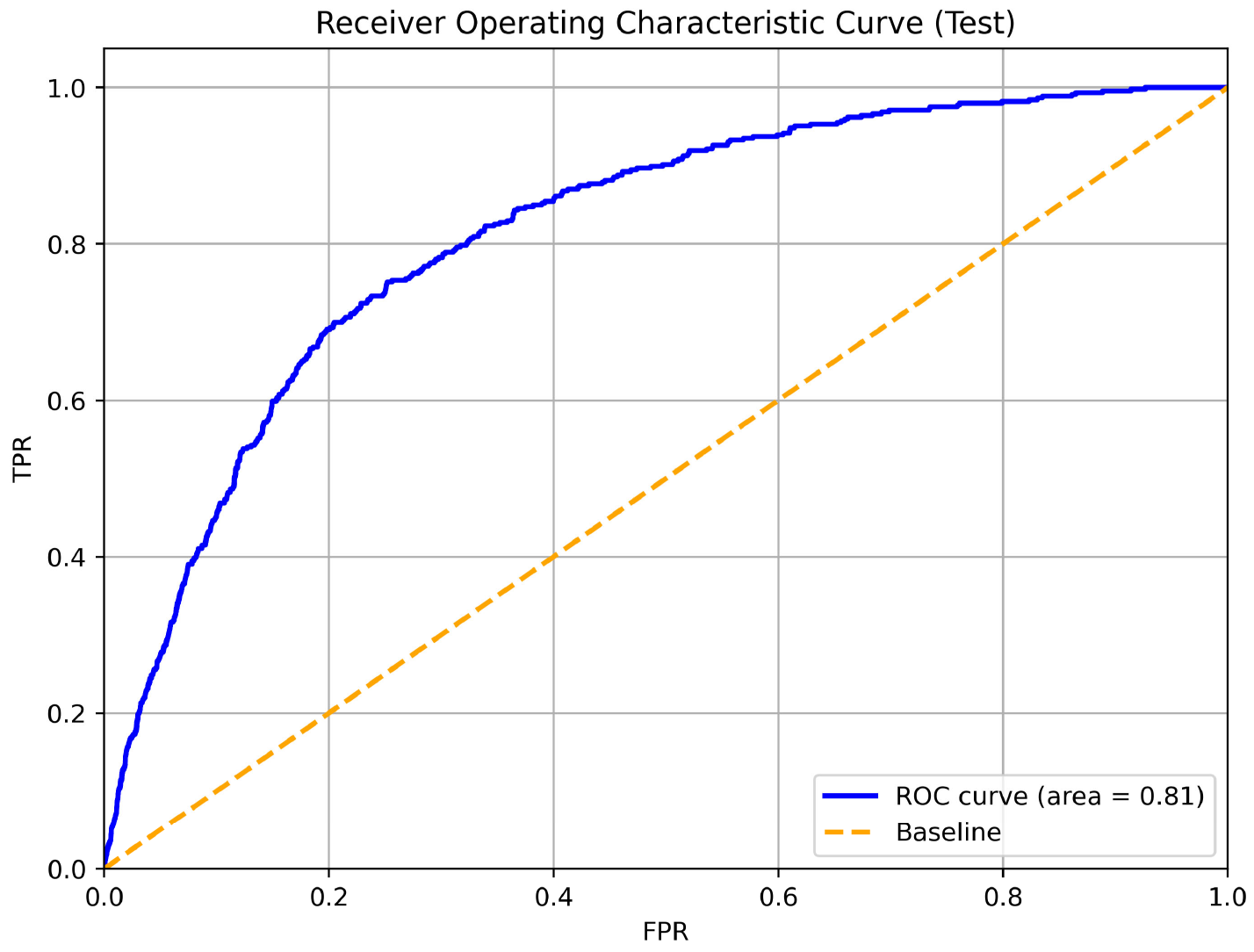}%
            \label{subfig:sepsis_pred_d}%
        } \hfill
        \subfloat[]{%
            \includegraphics[width=\twidth\linewidth]{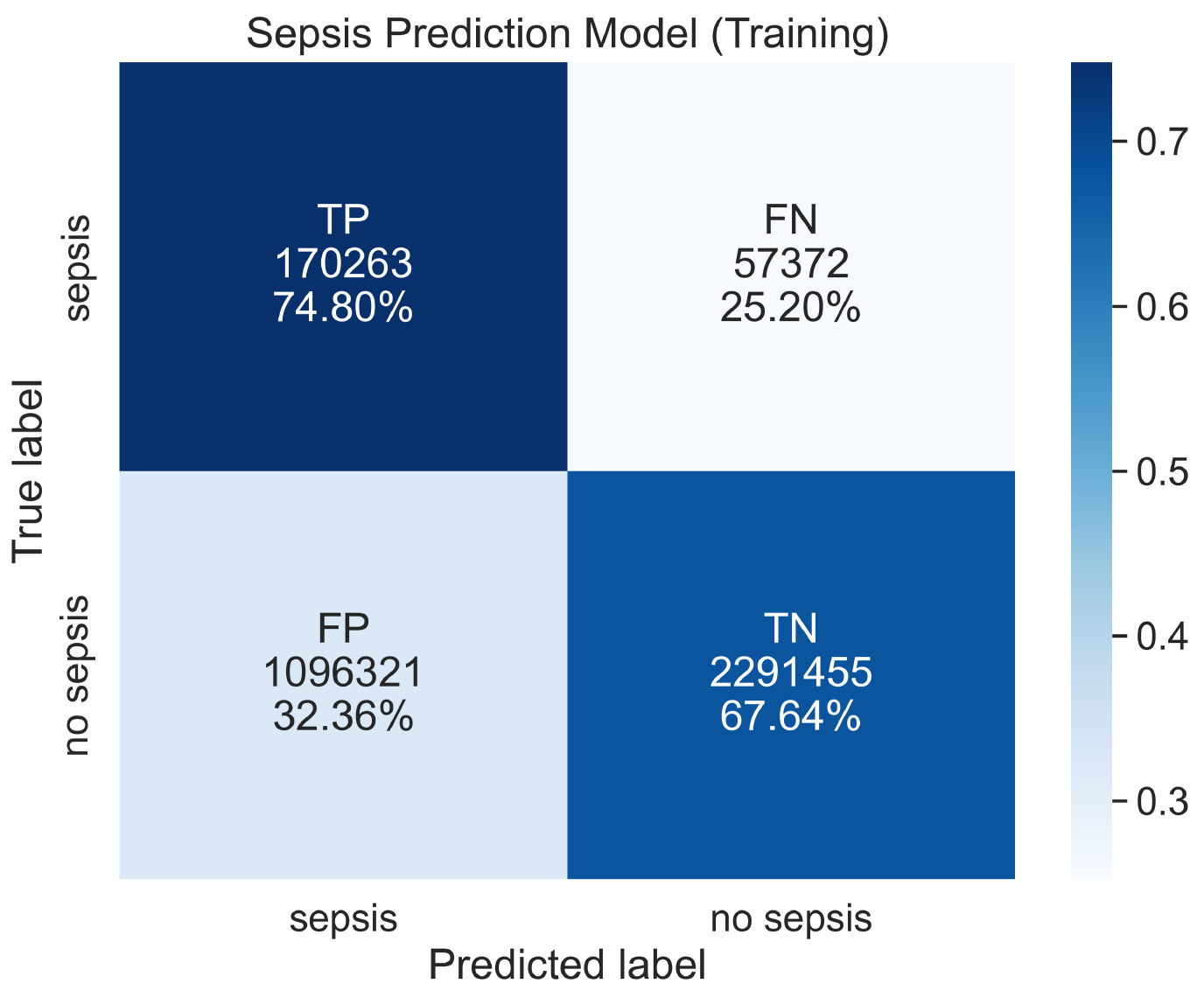}%
            \label{subfig:emory_sepsis_pred_a}%
        }
        \subfloat[]{%
            \includegraphics[width=\twidth\linewidth]{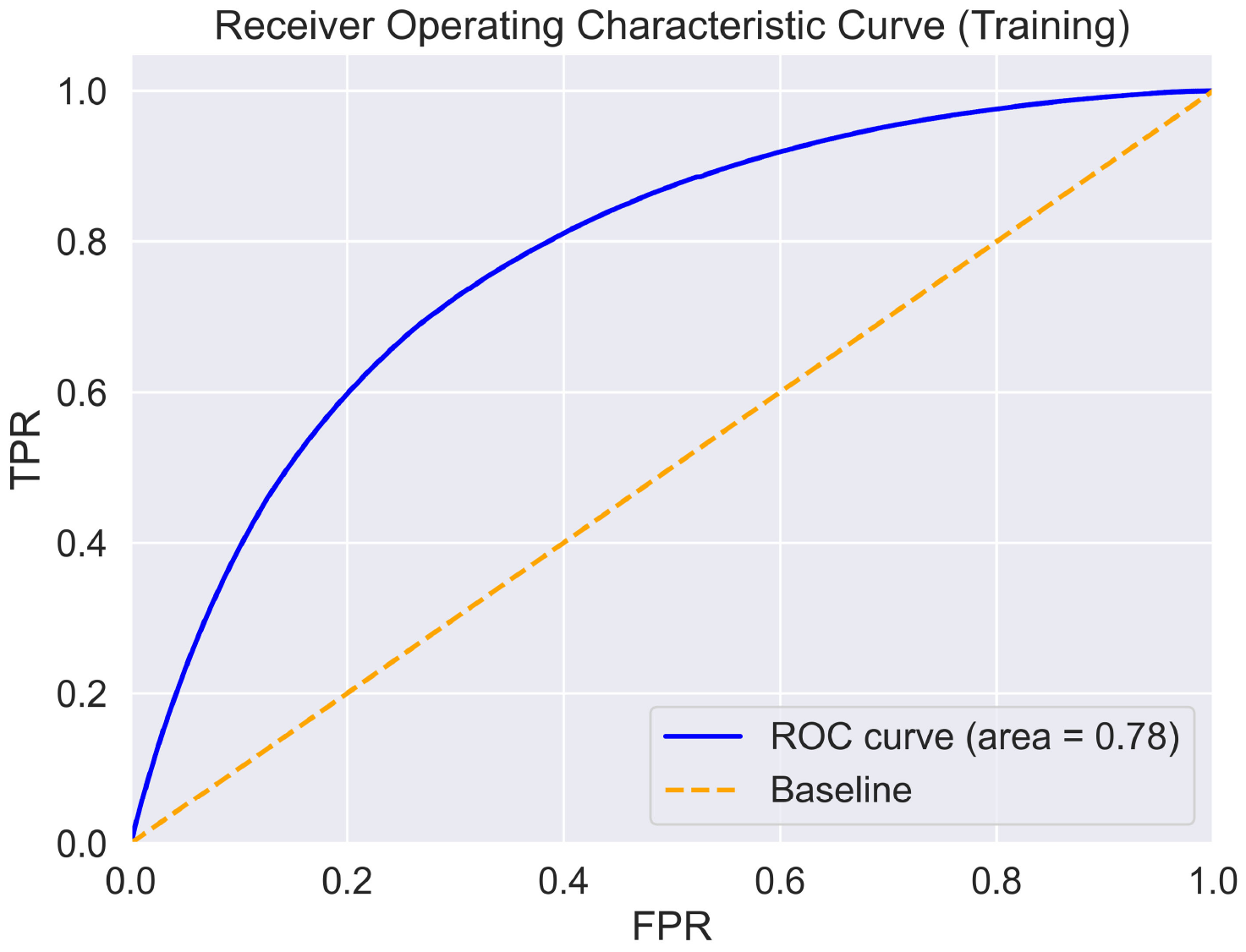}%
            \label{subfig:emory_sepsis_pred_b}%
        }
        \subfloat[]{%
            \includegraphics[width=\twidth\linewidth]{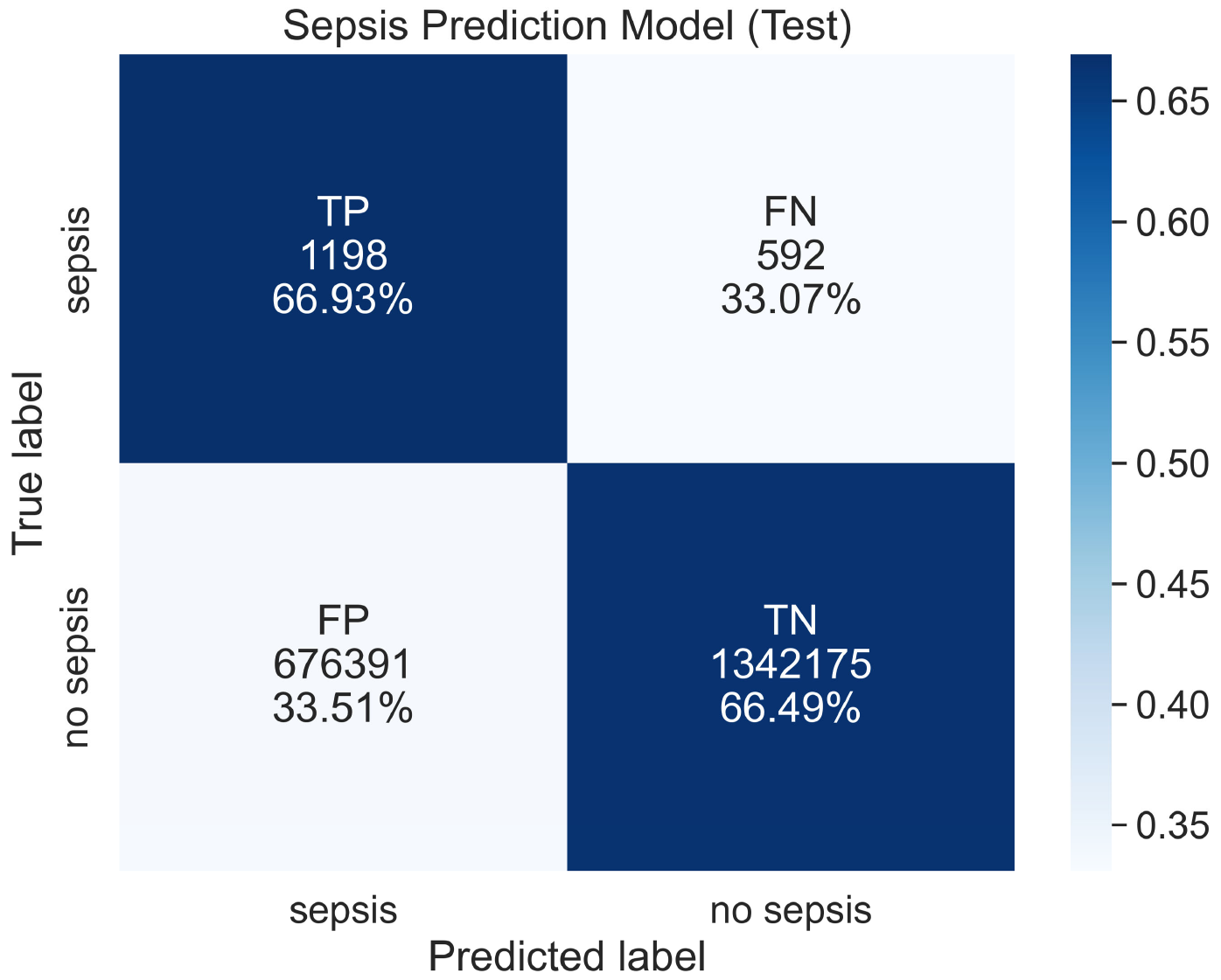}%
            \label{subfig:emory_sepsis_pred_c}%
        }
        \subfloat[]{%
            \includegraphics[width=\twidth\linewidth]{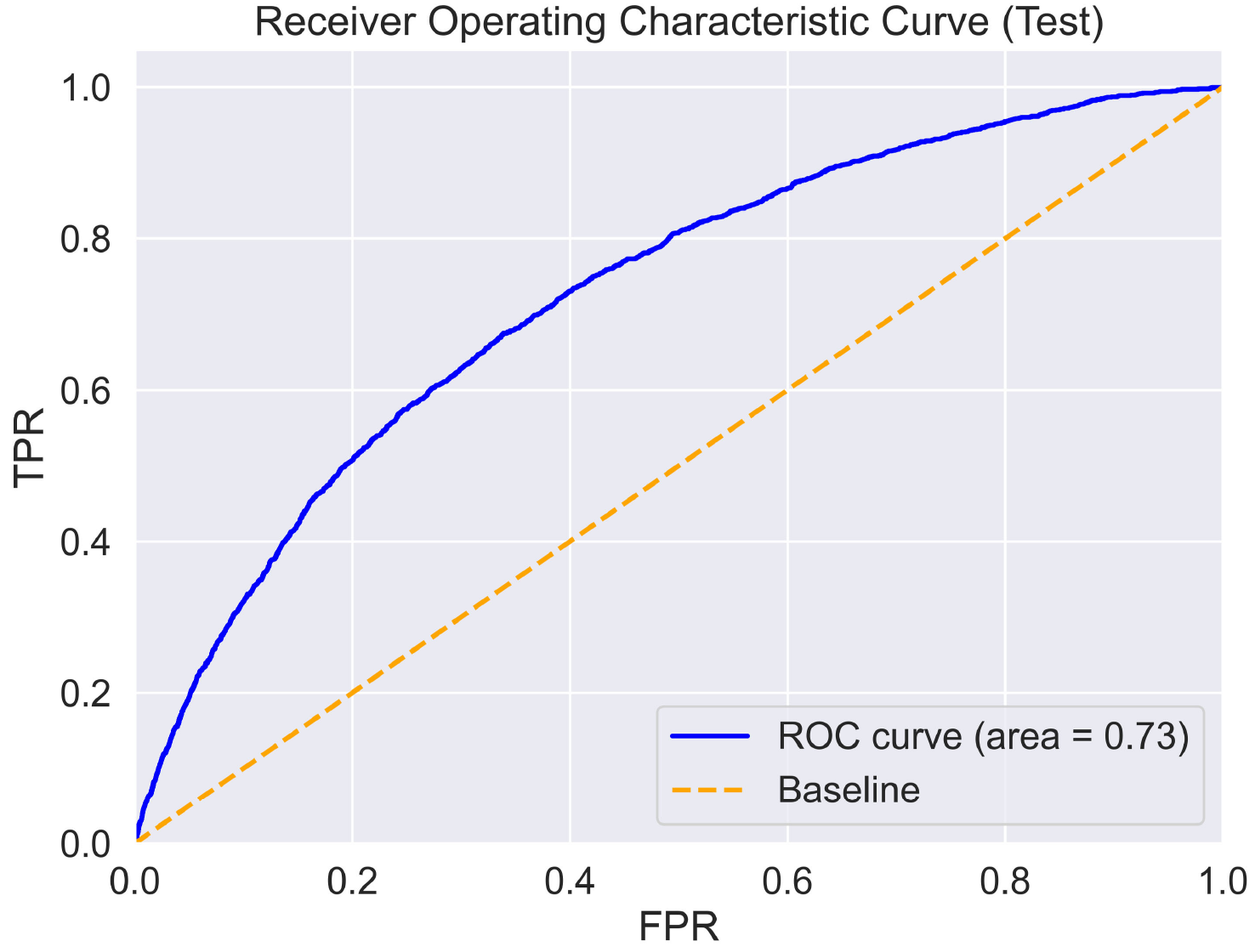}%
            \label{subfig:emory_sepsis_pred_d}%
            }
    \caption{The plots present the sepsis prediction model's performance measures. Plots (a) and (b) show the confusion matrix and ROC curve results of the model against the training data, respectively. Plots (c) and (d) provide similar measures for the test dataset.}
    \label{fig:sepsis_pred}
\end{figure}

\begin{table}[ht!]
\centering
\caption{Performance of different models on local test set formed by ourselves}
\begin{tabular}{c|c|c|c|c}
\hline
& \multicolumn{2}{c|}{{Grady}}& \multicolumn{2}{c}{{Emory}}\\
{XGBoost Models (Folds)} & {Accuracy} & {AUC} & {Accuracy} & {AUC}
 \\ \hline
1 & 0.840 & 0.728 & 0.637 & 0.643  \\ \hline
2 & 0.843 & 0.732 & 0.640 & 0.648  \\ \hline
3 & 0.791 & 0.712 & 0.670 & 0.629 \\ \hline
4 & 0.790 & 0.711 & 0.602 & 0.643 \\ \hline
5 & 0.790 & 0.712 & 0.691 & 0.647 \\ \hline
{Average} & \textbf{0.814} & \textbf{0.722} & \textbf{0.651} & \textbf{0.646} \\ \hline
{Ensemble Model} & \textbf{0.824} & \textbf{0.738} & \textbf{0.665} & \textbf{0.667}\\ \hline
\end{tabular}
    \label{fig:ens_model_results}
\end{table}
Fig. \ref{fig:sepsis_pred} provides a comprehensive visualization of the sepsis prediction models' performance for both Grady and Emory cohorts, across multiple evaluation metrics. The first row represents results from the model trained on Grady data, while the second row corresponds to the model trained on Emory data. These results are further categorized by the training and testing phases of model development. Figs. \ref{subfig:sepsis_pred_a} and \ref{subfig:emory_sepsis_pred_a} depict confusion matrices based on the respective training datasets. The receiver operator characteristics (ROC) curves, shown in Figs. \ref{subfig:sepsis_pred_b} and \ref{subfig:emory_sepsis_pred_b}, evaluate the model's ability to generalize to unseen test data. Figs. \ref{subfig:sepsis_pred_c} and \ref{subfig:emory_sepsis_pred_c} present confusion matrices for the test datasets, highlighting each model's predictive accuracy on unseen data. Finally, Figs. \ref{subfig:sepsis_pred_d} and \ref{subfig:emory_sepsis_pred_d} display the ROC curves for the test data. Table \ref{tab:prediction_results} provides a detailed summary of the classification performance metrics across both cohorts, providing further insights into the accuracy, precision, recall, F1-score, and F2-score for each model.
%
%
        
%
\begin{table}[!htbp]
    \centering
    \begin{tabular}{c|lccccc}
        \hline
        \multirow{3}{*}{Grady} & & \underline{Accuracy} & \underline{Precision} & \underline{Recall} & \underline{F1-Score} & \underline{F2-Score} \\
        & Training Set & 0.807 & 0.004 & 0.743 & 0.009 & 0.022 \\
        & Test Set & 0.828 & 0.004 & 0.641 & 0.007 & 0.017 \\
        \hline
        \multirow{3}{*}{Emory} & & \underline{Accuracy} & \underline{Precision} & \underline{Recall} & \underline{F1-Score} & \underline{F2-Score} \\
        & Training Set & 0.806 & 0.004 & 0.736 & 0.009 & 0.022 \\
        & Test Set & 0.824 & 0.003 & 0.652 & 0.007 & 0.017 \\
        \hline
    \end{tabular}
    
    \caption{Grady and Emory sepsis prediction model classification performance metrics for training and test sets.}
    \label{tab:prediction_results}
\end{table}

\end{document}